\documentclass{article}

\usepackage{arxiv}

\usepackage[utf8]{inputenc} 
\usepackage[T1]{fontenc}   
\usepackage{hyperref}     
\usepackage{url}           
\usepackage{booktabs} 
\usepackage{amsmath,amsfonts}
\usepackage{algorithmic}
\usepackage{algorithm}
\usepackage{array}    
\usepackage{nicefrac}
\usepackage{microtype}
\usepackage{graphicx}
\usepackage{caption}
\usepackage{natbib}
\usepackage{doi}
\usepackage{multirow,multicol}
\usepackage{pifont}
\usepackage[dvipsnames]{xcolor}
\usepackage{tabularx,booktabs,makecell}
\usepackage{wrapfig}

\renewcommand{\shorttitle}{}

\definecolor{SkyBlueCB}{HTML}{0072B2}

\title{Performance Optimization of YOLO-FEDER FusionNet for Robust Drone Detection in Visually Complex Environments}

\author{Tamara R. Lenhard$^{1,2,3}$ \href{https://orcid.org/0000-0001-9191-0170}{\includegraphics[scale=0.1]{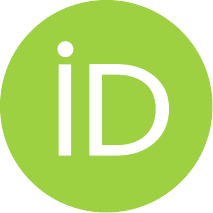}}\\
	\And
	Andreas Weinmann$^{2,3}$ \href{https://orcid.org/0000-0002-4969-7609}{\includegraphics[scale=0.1]{orcid.pdf}} \\
	\And
	Tobias Koch$^{1}$ \href{https://orcid.org/0000-0003-1279-0209}{\includegraphics[scale=0.1]{orcid.pdf}}\\
	\AND
	\\
	$^1$~{\small Institute for the Protection of Terrestrial Infrastructures, German Aerospace Center, Sankt Augustin, Germany}\\[2pt]
	$^2$~{\small ACIDA Lab, University of Applied Sciences Darmstadt, Darmstadt, Germany}\\[2pt]
	$^3$~{\small European University of Technology, European Union}\\[10pt]
	\texttt{\{tamara.lenhard, tobias.koch\}@dlr.de}\hspace{0.5cm}\texttt{andreas.weinmann@h-da.de} \\
}

\hypersetup{
pdftitle={A template for the arxiv style},
pdfsubject={q-bio.NC, q-bio.QM},
pdfauthor={Tamara R.~Lenhard, Andreas Weinmann, Tobias Koch},
pdfkeywords={drone detection, camouflage object detection, feature fusion, YOLO-FEDER FusionNet, synthetic data},
}

\begin{document}
\maketitle

\begin{abstract}
	Drone detection in visually complex environments remains challenging due to background clutter, small object scale, and camouflage effects. While generic object detectors like YOLO exhibit strong performance in low-texture scenes, their effectiveness degrades in cluttered environments with low object-background separability. To address these limitations, this work presents an enhanced iteration of YOLO-FEDER FusionNet -- a detection framework that integrates generic object detection with camouflage object detection techniques. Building upon the original architecture, the proposed iteration introduces systematic advancements in training data composition, feature fusion strategies, and backbone design. Specifically, the training process leverages large-scale, photo-realistic synthetic data, complemented by a small set of real-world samples, to enhance robustness under visually complex conditions. The contribution of intermediate multi-scale FEDER features is systematically evaluated, and detection performance is comprehensively benchmarked across multiple YOLO-based backbone configurations. Empirical results indicate that integrating intermediate FEDER features, in combination with backbone upgrades, contributes to notable performance improvements. In the most promising configuration -- YOLO-FEDER FusionNet with a YOLOv8l backbone and FEDER features derived from the DWD module -- these enhancements lead to a FNR reduction of up to 39.1 percentage points and a mAP increase of up to 62.8 percentage points at an IoU threshold of 0.5, compared to the initial baseline.
\end{abstract}

% keywords can be removed
\keywords{Drone Detection \and Camouflage Object Detection \and Feature Fusion \and YOLO-FEDER FusionNet}

\section{Introduction}
The development of robust and reliable drone detection technologies has become essential for strengthening the security of infrastructures, protecting individual privacy, and maintaining regulatory compliance~\cite{Chiper:2022,Mohsan:2023}. Among prevalent detection strategies, image-based drone detection has emerged as a highly promising and effective approach, exhibiting considerable potential for practical security applications. The growing adoption of image-based detection techniques is primarily driven by the economic viability, widespread availability, and inherent compatibility of camera sensors with existing surveillance infrastructures~\cite{Elsayed:2021}. By employing advanced computer vision algorithms, these techniques facilitate the timely identification of potential aerial threats and the implementation of effective countermeasures.

\begin{figure}
\centering
  \includegraphics[width=0.65\textwidth, trim={0.1cm 16.2cm 0.1cm 0cm}, clip]{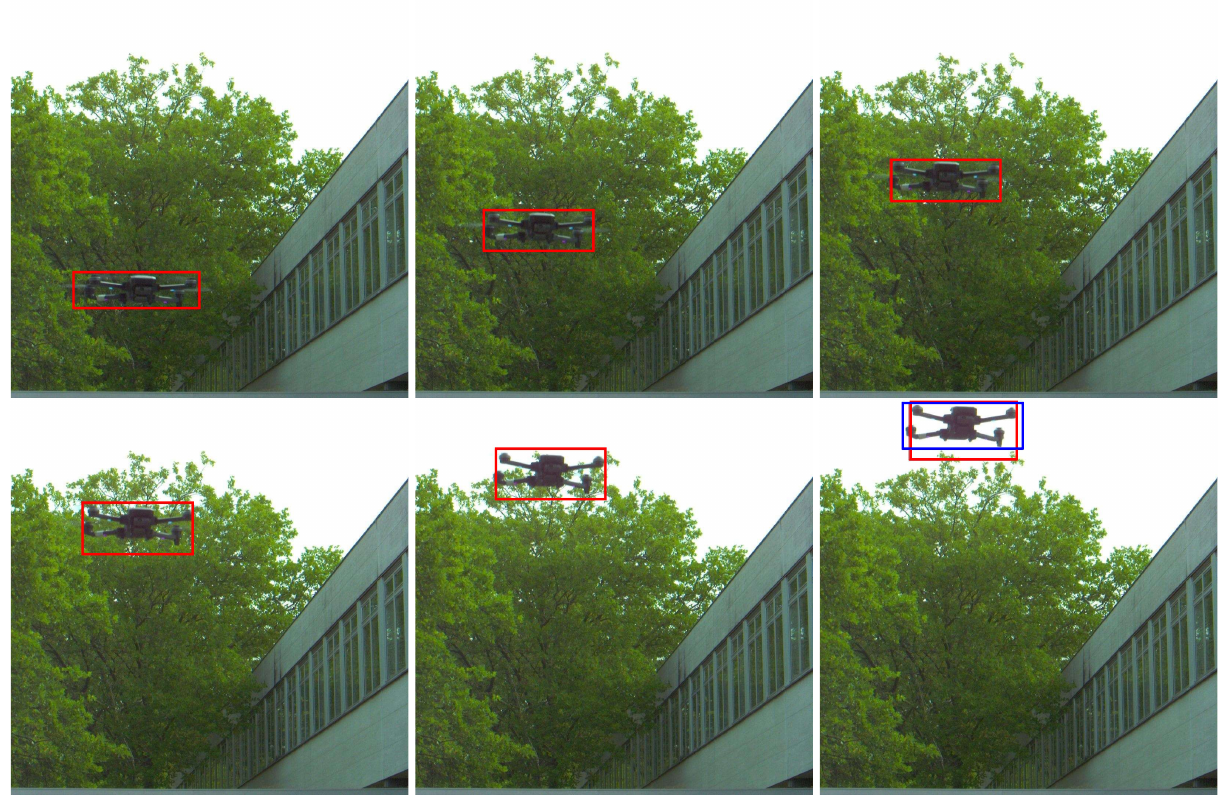}
  \caption{\label{fig:Comparison} Detection results across every fifth frame. While YOLO-FEDER FusionNet (red) detects the drone in all frames, YOLOv5l (blue) only detects it in the final frame. (Lenhard et al.~\cite{Lenhard:2024_YOLO-FEDER}, \copyright 2025 IEEE)}
\end{figure}

In drone detection applications, image analysis is typically performed using generic object detection models (e.g., YOLOv5 or YOLOv8~\cite{Ultralytics}). The broad adoption of these models is largely driven by their ability to achieve an optimal balance between real-time processing speed and detection accuracy. Moreover, generic object detection models have shown strong performance in scenarios where drones are positioned against uniform or low-clutter backgrounds -- such as clear blue skies~-- or in environments characterized by high visual contrast between the drone and its surroundings~\cite{Seidaliyeva:2020}. However, they often exhibit poor performance in visually complex scenarios featuring highly textured backgrounds~\cite{Seidaliyeva:2020,Dieter:2023,Liu:2024}. In previous studies, we highlighted the considerable challenges associated with detecting drones in densely vegetated environments~\cite{Dieter:2023}. The irregular structure of vegetation-covered backgrounds, combined with complex and dynamic lighting conditions, significantly impairs the detectability of drones (see Fig.~\ref{fig:Comparison}). Furthermore, the visual similarity between drone components -- such as rotor arms -- and adjacent branches further hinders reliable discrimination and facilitates their visual integration into the surrounding environment. Consequently, camouflage effects significantly degrade the performance of generic object detection models, adversely affecting both their accuracy and operational robustness~\cite{Dieter:2023}. Beyond drone detection, camouflage effects also pose substantial challenges in other domains -- particularly in animal detection -- where they have led the development of specialized \textit{camouflage object detection (COD)} techniques~\cite{Fan:2020}.

Despite their effectiveness in animal detection, the application of COD techniques to drone detection remains largely unexplored. To address this gap, we previously introduced a first version of \textit{YOLO-FEDER FusionNet} in the conference proceeding~\cite{Lenhard:2024_YOLO-FEDER}. YOLO-FEDER FusionNet is a novel deep learning (DL) architecture that combines generic object detection with the specialized capabilities of COD. Specifically, it integrates YOLO for generic feature encoding and FEDER~\cite{He:2023} for COD-specific feature representation within a unified network architecture. Building upon~\cite{Lenhard:2024_YOLO-FEDER}, this study presents an enhanced version of the original YOLO-FEDER FusionNet architecture and introduces the following key \textbf{contributions}:\vspace{-0.3cm}
\begin{itemize}
    \item \textit{Training Data Optimization:} To improve the detection capabilities of YOLO-FEDER FusionNet, we strategically optimize the underlying training data by integrating large-scale, high-fidelity synthetic RGB images with a small share of real-world samples. We demonstrate the effectiveness of the optimized dataset by highlighting its performance gains over the initial YOLO-FEDER FusionNet training strategy~\cite{Lenhard:2024_YOLO-FEDER}.
    \item \textit{Impact Analysis of Intermediate FEDER Features:} While the YOLO-FEDER FusionNet version in~\cite{Lenhard:2024_YOLO-FEDER} exclusively leverages the final output of the FEDER module -- specifically, the binary segmentation mask -- the potential contribution of FEDER's intermediate feature representations remains unexplored. Therefore, we conduct a comprehensive analysis on the integration of diverse hierarchical FEDER features within YOLO-FEDER FusionNet to assess their informational value and quantify their impact on detection performance.
    \item \textit{Evaluation of YOLO-Based Backbone Variations:} We investigate the architectural adaptability of YOLO-FEDER FusionNet by substituting its original generic feature encoder (YOLOv5l) with more advanced object detection architectures, including YOLOv8, YOLOv9, and YOLOv11. A comparative analysis is performed to evaluate the impact of these substitutions on detection accuracy and computational efficiency.
    \item \textit{Detection Error Assessment:} To systematically assess current performance constraints and identify opportunities for further refinement of YOLO-FEDER FusionNet, we perform a comprehensive evaluation of detection inaccuracies, explicitly focusing on the analysis and characterization of false-positive detections generated by the network.
\end{itemize}

The paper is organized as follows: Sec.~\ref{sec:relatedwork} provides an overview of recent advancements and relevant state-of-the-art techniques. Sect.~\ref{sec:framework} provides a detailed exposition of the architectural design of YOLO-FEDER FusionNet. The experimental setup -- including datasets, pre-processing strategies, training and evaluation configurations, as well as corresponding experimental procedures -- is outlined in Sec.~\ref{sec:expsetup}. Experimental results are presented and discussed in Sec.~\ref{sec:results}, followed by conclusions in Sec.~\ref{sec:conclusion}. Additional details are provided in the Supp. Material (Secs.~\ref{supp:module_comparison}--\ref{supp:performance_evolution}).

% RELATED WORK
\section{\label{sec:relatedwork}Related Work}
This section offers a comprehensive overview of relevant research across three key domains: image-based drone detection (Sec.~\ref{subsec:dronedetect}), camouflage object detection (Sec.~\ref{subsec:cod}), and the simulation-reality gap (Sec.~\ref{subsec:sim2realgap}).

\subsection{\label{subsec:dronedetect}Drone Detection}
Achieving reliable image-based drone detection typically requires addressing a diverse set of (interconnected) challenges. These include the accurate identification of small drones~\cite{Lv:2022,Liu:2021,Chen:2024,Kim:2023,Dong:2023}, their differentiation from visually similar aerial entities such as birds~\cite{Lv:2022}, and maintaining robust detection performance in visually complex, highly textured, or dynamically changing environments~\cite{Lv:2022,Dieter:2023,Lenhard:2024_YOLO-FEDER}. Despite advancements in small drone identification and aerial object differentiation~\cite{Lv:2022,Liu:2021},  the explicit development of techniques to ensure robust performance under complex visual conditions remains limited~\cite{Lv:2022,Dieter:2023}. Recent strategies built on RGB data are primarily focused on enhancing detection performance in complex scenes by systematically refining individual components of established object detection models~\cite{Lv:2022}. For instance, Lv et al.~\cite{Lv:2022} introduced a customized version of YOLOv5s, called SAG-YOLOv5s. The proposed model incorporates SimAM attention~\cite{Yang:2021} and Ghost modules~\cite{Han:2020} into the bottleneck layers of YOLOv5s to improve target-specific feature representation and suppress irrelevant information during feature extraction. 

Other techniques address scene complexity by adopting methodologically distinct strategies. One common approach involves decomposing the detection process into two successive stages: background elimination~\cite{Seidaliyeva:2020,Liu:2024} and moving object extraction~\cite{Chen:2017}, followed by classification. Alternatively, image-based detection techniques are employed either as initial stages within object tracking pipelines or as integral components of multi-sensor fusion frameworks~\cite{Svanstroem:2022}. This strategic integration is intended to compensate for potential shortcomings of purely camera-based systems, enhancing overall robustness -- particularly in visually complex or dynamic environments. 

However, the challenge of camouflage effects induced by natural surroundings -- such as vegetation and tree canopies -- has not been explicitly addressed in existing methodologies (apart from our prior work in~\cite{Lenhard:2024_YOLO-FEDER}).

\subsection{\label{subsec:cod}Camouflage Object Detection}
Camouflage object detection (COD) is an emerging research domain focused on developing advanced methodologies for identifying objects that closely replicate the intrinsic characteristics of their surroundings~\cite{Fan:2020}. These techniques are specifically designed to handle scenarios where objects exhibit minimal visual contrast and ambiguous boundaries, making accurate detection particularly challenging. Most existing approaches seek to overcome these challenges by modeling human visual perception~\cite{Fan:2020,Jia:2022}. Only a limited subset of COD techniques takes an alternative approach by disassembling the camouflage scenario and emphasizing subtle distinguishing features~\cite{He:2023}. A promising contribution within this category is the feature decomposition and edge reconstruction (FEDER) model by He et al.~\cite{He:2023}. In the context of animal detection, COD techniques have demonstrated notable effectiveness in addressing the challenges posed by naturally occurring camouflage.

\begin{figure}[!t]
\centering
\includegraphics[width=0.9\textwidth, trim={1.2cm 21cm 1.2cm 0.1cm}, clip]{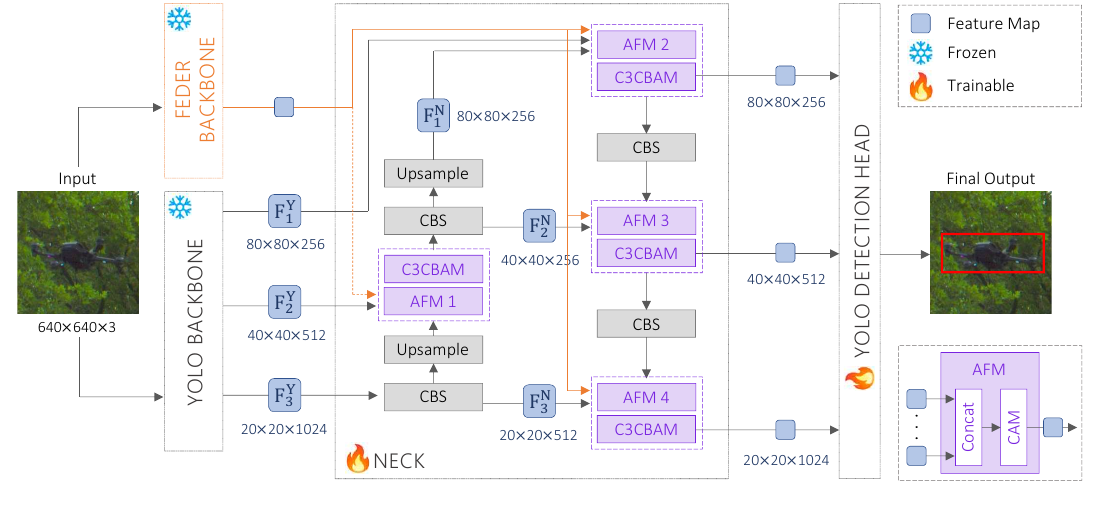}
\caption{\label{fig:yolofeder} Model architecture of YOLO-FEDER FusionNet. Primary components responsible for integrating and processing features from both backbones are marked in purple. Layer abbreviations include: AFM (attention fusion module), CAM (channel attention module), CBS (convolution, batch normalization, and SiLU activation), and C3CBAM (CSP bottleneck with three convolutional layers followed by channel attention). The feature maps $F^Y_i$, $i\in\{1,2,3\}$ are derived from the YOLO backbone, while the feature maps $F^N_i$ are obtained from neck components.}

\end{figure}

\subsection{\label{subsec:sim2realgap}Simulation-Reality Gap}
As real-world data collection remains resource-intensive and inherently limited in diversity, leveraging synthetically generated data has become a crucial strategy for training deep learning (DL) models in drone detection~\cite{Barisic:2022,Marez:2020,Symeonidis:2022,Dieter:2023} and other application domains~\cite{Pham:2022}. The synthesis of comprehensive, application-specific datasets is typically achieved by employing physically realistic, game engine-based simulations~\cite{Dieter:2022} or advanced domain randomization techniques~\cite{Marez:2020}. Compared to manual labeling processes, automated annotation techniques offer superior pixel-level accuracy, substantially lower associated costs, and mitigate inherent biases present in human-generated annotations. Moreover, these techniques enhance data diversity by overcoming real-world constraints, such as adverse weather conditions and privacy regulations. 

Despite the benefits, the transition of synthetic-data-trained models to real-world applications remains challenging due to the simulation-reality gap. This discrepancy can lead to performance degradation, the severity of which is primarily determined by the quality and representativeness of both synthetic and real-world data. The gap is commonly measured using performance indicators such as \textit{mean average precision (mAP)} across varying \textit{intersection over union (IoU)} thresholds~\cite{Barisic:2022,Marez:2020}. Common strategies for mitigating the simulation-reality gap are mixed-data training~\cite{Chen:2017} and fine-tuning with real-world data~\cite{Marez:2020}.

% FRAMEWORK
\section{\label{sec:framework}Framework}
YOLO-FEDER FusionNet effectively incorporates the capabilities of generic object detection algorithms with the distinctive strengths of camouflage object detection techniques. To achieve reliable detection performance, YOLO-FEDER FusionNet leverages two backbone components for extracting meaningful features: the state-of-the-art object detection model, YOLO~\cite{Ultralytics}, and the camouflage object detector, FEDER~\cite{He:2023}. Given their distinct focus and complementary outcomes, both backbone architectures function as an ensemble system. Thus, an input image $\mathbf{X}\in \mathbb{R}^{W\times H\times 3}$ is processed simultaneously by both backbone components (cf. Fig.~\ref{fig:yolofeder}). The extracted features are fused and refined within the network's neck, which is designed based on the architectural principles of YOLO~\cite{Ultralytics}. The refined feature maps are further processed by the network's detection head to generate (anchor-free) predictions at three distinct scales. A comprehensive overview of all YOLO-FEDER FusionNet components is provided in the subsequent sections.

\begin{table}[t!]
  \centering
  \footnotesize
  \caption{\label{tab:yolobckbnspecs}Technical overview of YOLO backbone architectures.}
 \begin{tabular}{lccccccccc}
 \hline\noalign{\smallskip}
 \multicolumn{1}{c}{YOLO Model} & No. Layers & No. Param (M). & GFLOPs (B)\\\noalign{\smallskip}\hline\noalign{\smallskip}
 \hspace{0.5cm}v5l  & 217 & 26.6 & 69.8 \\\noalign{\smallskip}\hline\noalign{\smallskip}
 \hspace{0.5cm}v8m  & 142 & 11.8 & 38.8 \\\noalign{\smallskip}\hline\noalign{\smallskip}
 \hspace{0.5cm}v8l  & 184 & 19.8 & 86.0 \\\noalign{\smallskip}\hline\noalign{\smallskip}
 \hspace{0.5cm}v9c  & 293 & 9.0 & 42.8 \\\noalign{\smallskip}\hline\noalign{\smallskip}
 \hspace{0.5cm}v9e  & 79  & 30.2 & 117.2 \\\noalign{\smallskip}\hline\noalign{\smallskip}
 \hspace{0.5cm}v11l & 238 & 12.7 & 48.0 \\\noalign{\smallskip}\hline\noalign{\smallskip}
 \hspace{0.5cm}v11x & 238 & 28.7 & 107.5 \\\hline
 \\
 \end{tabular}
\end{table}

\subsection{\label{subsec:yolobckbn}YOLO Backbone}
A variety of pre-trained YOLO backbone architectures are employed to facilitate the extraction of generic features. Specifically, this study leverages the backbone architectures of YOLOv5, YOLOv8, YOLOv9, and YOLOv11~\cite{Ultralytics}. These backbones build upon the architectural design of CSPDarkNet53, which combines DarkNet-53~\cite{Redmon:2018} with an advanced cross stage partial network (CSPNet) strategy~\cite{Wang:2020}. The latest YOLO iterations -- specifically YOLOv5, YOLOv8, and YOLOv11 -- are defined by a sequential arrangement of CBS modules, comprising convolutional layers, batch normalization, and SiLU activation functions. While variations in architectural depth and trainable parameters exist (see Tab.~\ref{tab:yolobckbnspecs}), the primary distinguishing features across these backbones are the specialized convolutional blocks: C3, C2f, and C3K2. A visual comparison of these blocks is provided in the Supp. Material (Sec.~\ref{supp:module_comparison}, Fig.~\ref{fig:ModuleComparison}). 

Further architectural distinctions are observed in the final layers of the backbones. While YOLOv5 and YOLOv8 terminate with a spatial pyramid pooling fusion (SPPF) module, YOLOv11’s architecture incorporates both SPPF and a C2PSA module, which embeds two partial spatial attention (PSA) mechanisms. In contrast, YOLOv9 adopts a markedly distinct backbone architecture, centered around an advanced variant of the CSP-ELAN module (RepNCSPELAN)~\cite{Wang:2024}. The RepNCSPELAN module consists of a repeated normalized cross-stage partial block (RepNCSP) combined with an efficient large kernel attention network (ELAN). Furthermore,  it employs an asymmetric downsampling technique and terminates with an SSPELAN module -- a shared spatial pyramid block integrated with ELAN~\cite{Wang:2023}.

Despite variations in convolutional block types and network depth, all backbone architectures consistently output three feature maps, denoted by $\mathbf{F}^Y_i$, $i \in \{1, \dots, 3\}$, with fixed dimensions: $20\times20\times1024$, $40\times40\times512$, and $80\times80\times256$ (cf. Fig.~\ref{fig:yolofeder}). These multi-scale feature maps are then processed by the neck architecture of YOLO-FEDER FusionNet for feature aggregation and refinement (see Sec.~\ref{subsec:neck}).

\subsection{\label{subsec:federbckbn}FEDER Backbone}
The feature decomposition and edge reconstruction (FEDER) model, developed by He et al.~\cite{He:2023}, comprises 1,186 layers, 44.1 million trainable parameters, and 200.1 billion GFLOPs. The model is built upon three key components: a camouflaged feature encoder (CFE), a deep wavelet-like decomposition (DWD) module, and a segmentation-oriented edge-assisted decoder (SED). Details of these components are provided in the following.\vspace{-0.1cm}

\begin{wrapfigure}{r}{0.5\linewidth} % r=right, l=left
  \vspace{-\intextsep}                 % pull image up to the paragraph
  \centering
  \includegraphics[width=0.5\textwidth, trim={0cm 18.2cm 11.6cm 0cm}, clip]{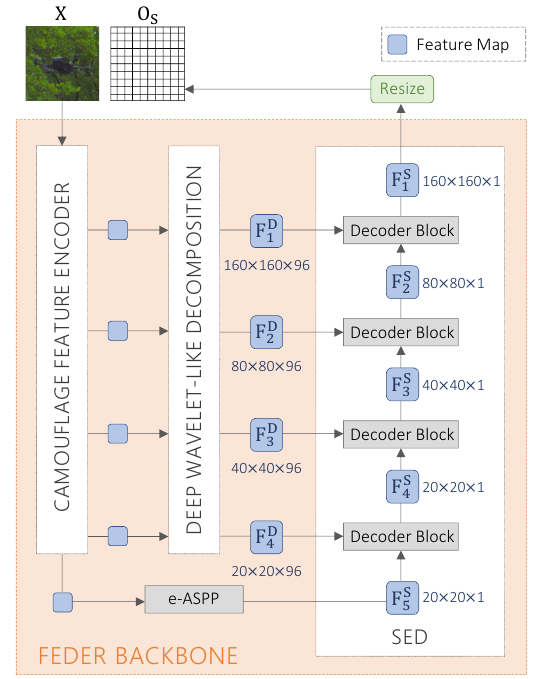}
  \caption{\label{fig:feder} Schematic representation of the FEDER backbone architecture~\cite{He:2023}, emphasizing the propagation of feature maps $F^D_i$, $i\in \{1,\dots,4\}$ and $F^S_j$, $j\in \{1,\dots,5\}$ across successive layers.}
\end{wrapfigure}

\paragraph{\hspace{-0.1cm}Camouflaged Feature Encoder (CFE).} The CFE integrates a Res2Net50 architecture~\cite{Gao:2019} alongside R-Net~\cite{Fan:2020} to extract multi-scale feature representations from an input image $\mathbf{X}\in \mathbb{R}^{W\times H\times 3}$, where $W=H$. The extracted 64-channel feature maps are further refined by the efficient atrous spatial pyramid pooling (e-ASPP)~\cite{Chen:2016} module for multi-scale contextual aggregation and the DWD module (cf. Fig.~\ref{fig:feder}).

\paragraph{\hspace{-0.1cm}Deep Wavelet-like Decomposition (DWD) Module.} In COD, discriminative features predominantly arise from high-frequency (HF) and low-frequency (LF) components~\cite{Stevens:2009}. While HF components encode fine-grained structural details, such as texture and edge variations, LF components encapsulate global attributes, including color distribution and illumination patterns. The DWD module exploits this inherent spectral distinction by applying learnable HF and LF filters, coupled with adaptive wavelet distillation~\cite{Ha:2021}, to systematically decompose the feature maps extracted from the CFE. By leveraging HF and LF attention modules in conjunction with a guidance-based feature aggregation technique, the DWD enhances the refinement of decomposed features, fosters inter-feature interaction, and ensures a semantically meaningful fusion strategy. The feature maps $F^D_i$, $i\in\{1,\dots,4\}$ extracted by the DWD module serve as the primary input for the segmentation-oriented edge-assisted decoder (see Fig.~\ref{fig:feder}).\vspace{-0.1cm}

\paragraph{\hspace{-0.1cm}Segmentation-oriented Edge-assisted Decoder (SED).} The SED framework consists of two principal components for refined feature learning and auxiliary edge reconstruction: a reversible re-calibration segmentation (RRS) module and an ordinary differential equation (ODE)-inspired edge reconstruction (OER) module. While the RRS module seeks to amplify discriminative features in low-confidence regions  via an inverse attention mechanism, the OER module focuses on enhancing edge-aware feature representation. Inspired by the dynamical systems perspective on deep neural networks (DNNs)~\cite{Weinan:2017}, the OER module builds on the interpretation of feature propagation in residual networks as discrete approximation of a non-linear ODE. Specifically, the evolution of feature states $\mathbf{y}_t$ across residual layers $t \in \{0, 1, \dots, T\}$ is described by
\begin{equation}
\label{eq:resblock}
    \mathbf{y}_{t+1} = \mathbf{y}_t+\Delta t~\sigma(\mathbf{K}(\mathbf{w}_t)\mathbf{y}_t+\mathbf{b}_t) \ , 
\end{equation}

\noindent which corresponds to the first-order (forward) Euler discretization of the continuous ODE:
\begin{equation}
\label{eq:ode}
    \frac{dy}{dt} = \sigma(w(t)\ast y(t)+b(t)), \ \ y(0)=\mathbf{L}x \ , \ \ t\in [0, T] \ .
\end{equation}

\noindent  Here, $\mathbf{K}(\mathbf{w}_t)$ denotes the convolutional matrix parameterized by learnable weights $\mathbf{w}_t$, $\mathbf{b}_t$ is the associated bias term, and $\sigma(\cdot)$ represents a non-linear activation function. The functions $w(t)$ and $b(t)$ in Eq.~\ref{eq:ode} correspond to the continuous analogs of $\mathbf{w}_t$ and $\mathbf{b}_t$, with $*$ denoting the continuous convolution operator. The learnable projection operator $\mathbf{L}$ maps the input $\mathbf{x}$ into the latent ODE state space. The step size $\Delta t > 0$ is fixed to one in standard DNN implementations.

To mitigate truncation errors inherent in first-order Euler discretization~\cite{He:2023,Weinan:2017}, the OER module adopts a second-order, explicit Runge-Kutta (RK)-inspired scheme for more accurate feature propagation. In contrast to the standard RK formulation, which requires a manually specified trade-off parameter, the OER module incorporates a learnable gating mechanism that adaptively modulates the contribution of intermediate feature states. To further enhance stability, the module also integrates propagation techniques inspired by Hamiltonian systems~\cite{Haber:2018}. 

The feature maps extracted by the SED are denoted by $\mathbf{F}^S_j$, $j\in\{1, \ldots, 4\}$ (see Fig.~\ref{fig:feder}). A comprehensive overview of the architectural design and mathematical formulation of both the OER and RSS modules can be found in~\cite{He:2023}.

\subsection{\label{subsec:neck}Neck}
YOLO-FEDER FusionNet employs a neck architecture designed to facilitate the integration of multi-scale feature representations from the two backbones across various hierarchical layers (see Fig.~\ref{fig:yolofeder}). Building upon the architectural principles of YOLOv5~\cite{Ultralytics}, the neck structure incorporates fundamental components such as CBS blocks, C3 modules, and upsampling layers. To enhance cross-backbone feature fusion, a modified concatenation module -- referred to as \textit{attention fusion module (AFM)} -- is introduced (cf. Fig.~\ref{fig:yolofeder}). The AFM combines feature concatenation with a subsequent channel attention mechanism designed to capture inter-channel dependencies within the aggregated feature space. The AFM-based fusion of, for instance, an intermediate YOLO feature map $\mathbf{F}^Y_i \in \mathbb{R}^{W \times H \times C}$, $i\in\{1,\dots,3\}$, and the FEDER-derived binary segmentation map $\mathbf{O}_S \in \mathbb{R}^{W \times H \times 1}$ can be formulated as follows:\vspace{0.1cm}
\begin{equation}
\label{eq:AFM}
\mathrm{AFM}(\mathbf{F}^Y_i, \mathbf{O}_S)  = \mathbf{M}_C(\mathbf{F}_C)\otimes \mathbf{F}_C \ .
\end{equation}

\noindent Here, $\mathbf{F}_C = \mathrm{Concat}(\mathbf{F}^Y_i, \mathbf{O}_S)$ denotes the concatenated feature map, while $\mathbf{M}_C(\mathbf{F}_C) \in \mathbb{R}^{1 \times 1 \times C}$ represents the corresponding channel attention map. The operator $\otimes$ denotes element-wise multiplication, where $\mathbf{M}_C(\mathbf{F}_C)$ is broadcast along the spatial dimensions to match the size of $\mathbf{F}_C$~\cite{Woo:2018}. This attention-based refinement is particularly effective for integrating information from heterogeneous sources, enabling the network to selectively emphasize the most informative features from each input. 

To further enhance feature representation, the neck architecture also includes \textit{convolutional block attention modules (CBAM)}. CBAM -- originally introduced by Woo et al.~\cite{Woo:2018} -- is a widely employed attention mechanism that concurrently captures both spatial and channel-wise feature dependencies. Building upon the integration strategy presented by Woo et al.~\cite{Woo:2018}, where CBAM is embedded within the residual blocks of ResNet50, a comparable approach is adopted in YOLO-FEDER FusionNet. In fact, CBAM is embedded within the CSP bottleneck of the C3 module (see Fig.~\ref{fig:yolofeder}, C3CBAM) to selectively emphasize informative regions and suppress less relevant features. Given an intermediate feature map $\mathbf{F} \in \mathbb{R}^{W \times H \times C}$ derived from a preceding CBS block the overall attention process initiated by CBAM can be described as follows:
\begin{equation}
    \mathrm{CBAM}(\mathbf{F}) = \mathbf{M}_S(\mathbf{M}_C(\mathbf{F})\otimes \mathbf{F}) \otimes (\mathbf{M}_C(\mathbf{F})\otimes \mathbf{F} \ .
\end{equation}

Here, $\mathbf{M}_S(\mathbf{F}) \in \mathbb{R}^{W \times H \times 1}$ denotes the two-dimensional spatial attention map which is broadcast along the channel dimension to enable element-wise multiplication $\otimes$ with the input tensor $\mathbf{F}$.

\subsection{\label{subsec:head}Head} 
YOLO-FEDER FusionNet adopts a detection head architecture inspired by YOLOv8, integrating two convolutional layers and a \textit{distribution focal loss (DFL)} module to enhance bounding box regression accuracy. The head operates in an anchor-free fashion, eliminating the need for pre-defined anchors and improving generalization to objects with diverse shapes and sizes. To support multi-scale detection, predictions are made at three spatial resolutions: 80$\times$80 for small objects, 40$\times$40 for medium objects, and 20$\times$20 for large objects (cf. Fig.~\ref{fig:yolofeder}).

% EXPERIMENTAL SETUP
\section{\label{sec:expsetup}Experimental Setup}
The following sections provide a detailed overview of the experimental setup, covering the training and validation data  (Sec.~\ref{subsec:data}), data pre-processing strategies  (Sec.~\ref{subsec:preprocessing}), training and evaluation specifications (Secs.~\ref{subsec:train_details} and~\ref{subsec:eval_details}), and a description of the experimental procedure (Sec.~\ref{subsec:exp_details}).

\subsection{\label{subsec:data}Data}
Given the limited availability of publicly accessible datasets for image-based drone detection, this study integrates multiple data sources to enhance both training and evaluation. Specifically, we utilize a combination of real-world datasets -- comprising self-collected and publicly available data -- and synthetically generated datasets. While synthetic data is leveraged exclusively for training, real-world data constitutes the primary evaluation benchmark. Both dataset types have been introduced in our previous works \cite{Dieter:2023} and \cite{Lenhard:2024}. Dataset statistics are summarized in Tab.~\ref{tab:data}, with further details provided in subsequent sections.\vspace{-0.1cm}

\begin{table*}[t!]
\centering
\footnotesize
\caption{\label{tab:data}Overview of employed synthetic and real-world datasets.}
  \begin{tabular}{lcccccccccc}
  \hline\noalign{\smallskip}
    Dataset & Type & Pub.  & \multicolumn{4}{c}{Image Count} & Resolution & \multicolumn{2}{c}{Camera Param.} & Drone \\
    & &  Avail. &train & val & test & total & (px) & pos. & focal length & Models \\\noalign{\smallskip}\hline\noalign{\smallskip}
    R1 & real & \ding{55} & 7,524 & 2,508 & 2,508 & 12,540 & 2040$\times$1086 & 2 & 8 mm & 1\\\noalign{\smallskip}\hline\noalign{\smallskip}
    R2 & real & \ding{55} & 3,834 & 1,279 & 1,278 & 6,391 & 2040$\times$1086 & 2 & 25 mm & 1\\\noalign{\smallskip}\hline\noalign{\smallskip}
    S1 & synth. & \ding{55} & 10,446 & 3,483 & 3,483 & 17,412 & 2040$\times$1080 & 5 & -- & 3\\\noalign{\smallskip}\hline\noalign{\smallskip}
    DUT Anti-UAV~\cite{Zhao:2022} & real & \ding{51} & 5,200 & 2,600 & 2,200 & 10,000 & \ \ \ --$^{\text{\ding{72}}}$ & \ \  --$^{\text{\ding{115}}}$ & -- & 35\\\noalign{\smallskip}\hline\noalign{\smallskip}
    
    SynDroneVision~\cite{Lenhard:2024} & synth. & \ding{51} & 131,238 & 8,800 & 4,000 & 140,038 & 2560$\times$1489 & 58 & -- & 13\\\noalign{\smallskip}\hline\noalign{\smallskip}
 
    \multicolumn{11}{l}{{\footnotesize\ding{72}}{\footnotesize \ No uniform image resolution, variations between 240$\times$160 to 5616$\times$3744 pixels.}}\\
    
    \multicolumn{11}{l}{{\footnotesize\ding{115}}{\footnotesize \ Variations included, but no explicit specifications.}}\\
  \end{tabular}
\end{table*}

\paragraph{\hspace{-0.1cm}Self-Captured Real-World Data.} A ground-mounted Basler acA200-165c camera system, featuring 25mm and 8mm lenses, is used for real-world data collection. This configuration facilitates the acquisition of two distinct camera field-of-view from each vantage point. To ensure realistic data collection, the designated recording site closely resembles the architectural and environmental characteristics of a potential urban deployment location for drone detection systems (refer to~\cite{Dieter:2023} for more detailed information). The recorded RGB images are organized into two distinct datasets, R1 and R2 (cf. Tab.~\ref{tab:data}). Dataset R1 is characterized by urban infrastructure, with buildings comprising the majority of the background. Conversely, R2 exhibits a higher concentration of complex, highly textured elements (predominantly trees), contributing to an increased visual complexity. Both datasets are captured at a uniform resolution of 2040$\times$1086 pixels. Data annotation is conducted manually using CVAT.\vspace{-0.1cm}

\paragraph{\hspace{-0.1cm}Publicly Available Real-World Data.} One of the most representative publicly accessible datasets for drone detection in a surveillance setup is the Dalian University of Technology (DUT) Anti-UAV dataset~\cite{Zhao:2022}. Designed for both drone detection and tracking, this dataset features a broad range of image resolutions (from 240$\times$160 to 5616$\times$3744 pixels) and encompasses diverse drone models, environments, lighting conditions, and weather scenarios (see~\cite{Zhao:2022} for details).\vspace{-0.1cm}

\paragraph{\hspace{-0.1cm}Synthetic Data.} The creation of synthetic training data relies on a game engine-based data generation pipeline. By leveraging the advanced capabilities of Unreal Engine (4.27 and 5.0, respectively)~\cite{UnrealEngine} and Colosseum~\cite{Colosseum} -- a successor of AirSim~\cite{Airsim} -- the pipeline enables the systematic acquisition of automatically annotated RGB images. The pipeline's technical specifications are comprehensively detailed in~\cite{Dieter:2022}. Using this pipeline, two distinct synthetic datasets are generated (cf. Tab.~\ref{tab:data}): S1, a smaller and simpler dataset, and SynDroneVision~\cite{Lenhard:2024}, a large-scale dataset characterized by a wide variety of drone models, environments, and illumination conditions. While the generation of SynDroneVision prioritizes scalability and data diversity to reflect real-world complexities, S1 is specifically designed to capture the essential attributes of R1 and R2. The data in S1 is provided at a resolution of 2040$\times$1080 pixels, whereas SynDroneVision features a higher resolution of 2560$\times$1489 pixels. In addition to drone images, both datasets include a small share of background images (7-8\%). For a detailed description of SynDroneVision, refer to \cite{Lenhard:2024}. For further information on Dataset S1, see \cite{Dieter:2023}.

\subsection{\label{subsec:preprocessing}Data Pre-processing}
A two-stage cropping methodology is employed to meet the model’s requirement for square input images. The initial stage involves a coarse cropping strategy, guided by the precise localization of the drone's ground truth bounding box and predefined cropping dimensions. This step ensures the drone's retention in the subsequent cropping phase, irrespective of the cropping parameters. To enhance data variability, the second stage leverages a random cropping mechanism using distinct dimensions: 640$\times$640 (default input size for YOLO), 1080$\times$1080, and a dynamically determined size based on the smallest image dimension. The dynamic adaptation of cropping sizes optimizes the preservation of informational content, while maintaining flexibility across varying image resolutions. 
This cropping strategy generates diverse dataset representations, each aligned with the specified cropping dimensions (see Tab.~\ref{tab:cropping_dims}).\\[2pt]

\subsection{\label{subsec:train_details}Training Specifications}
YOLO-FEDER FusionNet is trained based on the following selection of augmentation techniques, hyperparameters, and a tailored loss function:\vspace{-0.1cm}

\begin{table}[t!]
\centering
\footnotesize
\caption{\label{tab:cropping_dims}Overview of datasets and corresponding cropping resolutions. The notation \textit{dyn.} indicates a dynamically adjusted cropping size.}
  \begin{tabular}{lcccc}
  \hline\noalign{\smallskip}
    Dataset & \multicolumn{3}{c}{Cropping Resolution} & Dataset\\
    & 640$\times$640 & 1080$\times$1080 & dyn.& Versions\\\noalign{\smallskip}\hline\noalign{\smallskip}
    R1 & \ding{51} & \ding{51} & -- & 2\\\noalign{\smallskip}\hline\noalign{\smallskip}
    R2 & \ding{51} & \ding{51} & -- & 2\\\noalign{\smallskip}\hline\noalign{\smallskip}
    S1 & \ding{51} & \ding{51} & -- & 1 \\\noalign{\smallskip}\hline\noalign{\smallskip}
    DUT Anti-UAV~\cite{Zhao:2022} & -- & -- & \ding{51} & 1\\\noalign{\smallskip}\hline\noalign{\smallskip}
    SynDroneVision~\cite{Lenhard:2024} & -- & -- & \ding{51} & 1 \\\noalign{\smallskip}\hline\noalign{\smallskip}
  \end{tabular}
\end{table}

\paragraph{\hspace{-0.1cm}Augmentation Techniques.} To enhance data diversity and improve model robustness, we employ a combination of standard augmentation techniques. This includes random horizontal flipping with a probability of 0.5 and HSV augmentation, where hue, saturation, and brightness are dynamically adjusted within normalized ranges of $\pm 0.015$, $\pm 0.7$, and $\pm 0.4$, respectively. Additionally, we incorporate a refined adaptation of mosaic augmentation, inspired by \cite{Ultralytics}. To mitigate biases in YOLO-based detection and prevent distortions in COD, we deliberately exclude letterboxing -- i.e., artificial padding used to maintain aspect ratios -- from mosaic augmentation. No additional blurring is applied, as the SynDroneVision dataset inherently includes blurred images, making further modifications unnecessary.\vspace{-0.1cm}

\paragraph{\hspace{-0.1cm}Hyperparameter Settings.} The training process is conducted over 100 epochs with a batch size of 64 and an input image size of 640$\times$640, utilizing four Nvidia A100 GPUs. Optimization is performed using stochastic gradient descent (SGD) with an initial learning rate of 0.01 and a momentum coefficient of 0.937. To further enhance convergence efficiency, an exponential moving average is applied. All other hyperparameter settings remain consistent with those used in~\cite{Ultralytics}.\vspace{-0.1cm}

\paragraph{\hspace{-0.1cm}Loss Function.} The loss function employed in the training process follows the structure of YOLOv8 and consists of three key components: the bounding box regression loss $\mathcal{L}_{\text{box}}$, the classification loss $\mathcal{L}_{\text{cls}}$, and the distribution focal loss $\mathcal{L}_{\text{dfl}}$. While $\mathcal{L}_{\text{cls}}$, formulated using binary cross-entropy (BCE), ensures precise and robust classification, both $\mathcal{L}_{\text{box}}$ and $\mathcal{L}_{\text{dfl}}$ contribute to bounding box localization. Specifically, the CIoU-based loss $\mathcal{L}_{\text{box}}$ enhances localization accuracy by optimizing spatial alignment, aspect ratio consistency, and proximity to ground truth. In contrast, $\mathcal{L}_{\text{dfl}}$ models a probability distribution over bounding box coordinates rather than directly regressing them. This is particularly effective  in scenarios involving both synthetic and real-world data, where the pixel-level precision of synthetic annotations contrasts with the inherent variability and uncertainty of manually annotated real-world data. The total loss $\mathcal{L}$ is expressed as
\begin{equation}
\label{eq:YOLOv5_loss}
\mathcal{L} = w_1\mathcal{L}_{\text{box}} + w_2\mathcal{L}_{\text{cls}} + w_3\mathcal{L}_{\text{dfl}} \ ,
\end{equation}

\noindent where the non-negative weighting coefficients $w_1$, $w_2$ and $w_3$ significantly influence the optimization dynamics and overall model performance. While YOLOv8’s default weight configuration is empirically tuned for the COCO benchmark dataset, these values may not generalize effectively across varying data domains. To ensure optimal performance for our specific application, we conducted a comprehensive ablation study, leading to the selection of $w_1=0.1$, $w_2=0.5$, and $w_3=1.5$. Details of the ablation study are provided in the Supp. Material (Sec.~\ref{supp:loss}, Tabs.~\ref{tab:bboox_loss_weight} and \ref{tab:bboox_loss_weight_avg_fitness}).

\subsection{\label{subsec:eval_details}Evaluation Details}
The evaluation of all trained iterations of YOLO-FEDER FusionNet (see Sec.~\ref{subsec:exp_details} for detailed specifications) is based upon the following datasets and performance indicators:\vspace{-0.2cm}

\paragraph{\hspace{-0.1cm}Data.} Performance evaluation relies exclusively on real-world data, utilizing the datasets R1 and R2 for both cropping dimensions (cf. Tabs.~\ref{tab:data} and \ref{tab:cropping_dims}).\vspace{-0.2cm}

\paragraph{\hspace{-0.1cm}Metrics.} The effective mitigation of unauthorized drone intrusions requires timely and accurate detection, making the reduction of false negatives -- measured by the false negative rate (FNR) -- a critical factor in ensuring system reliability. However, in practical surveillance applications, detecting a drone in every individual frame of an image sequence is not imperative. Undetected instances can often be inferred from adjacent frames, enabling a partial reconstruction of the drone’s presence.

Considering drone detection as a key component of a broader security framework, minimizing false positives -- quantified by the false discovery rate (FDR) -- is nearly as important as reducing false negatives to maintain system credibility and operational effectiveness. To provide a holistic evaluation of detection performance, we assess YOLO-FEDER FusionNet using FNR and FDR, alongside mAP at an IoU threshold of 0.5 (a widely adopted metric in object detection). Given that precise bounding box localization is less critical in our context and manually generated annotations often exhibit variability, we also consider mAP at an IoU threshold of 0.25.

To integrate these performance indicators into a single evaluation criterion, we introduce \textit{model fitness} (inspired by~\cite{Ultralytics,Silva:2023}), defined as a weighted sum of all key quality measures:\vspace{0.1cm}
\begin{equation}
\label{eq:fitness}
\text{fitness} = 0.45 \times (1-\text{FNR}) + 0.35 \times (1-\text{FDR}) + 0.10  \times \text{mAP}_{0.25} + 0.10  \times \text{mAP}_{0.5} \ .
\end{equation}

\noindent The weighting of individual components is defined according to the aforementioned prioritization. Nevertheless, empirical evaluations demonstrate that variations in the exact weight values have negligible impact, as long as their relative prioritization is maintained (cf. Sec.~\ref{supp:fitness}, Supp. Material).

\subsection{\label{subsec:exp_details}Experimental Procedure}
To assess the effectiveness of the proposed enhancements to YOLO-FEDER FusionNet, we conduct four structured experiments, each addressing a distinct focus area: training data optimization, feature integration, architectural modifications, and false positive mitigation. To ensure comparability, a consistent training configuration is applied across all DL models (see Sec.~\ref{subsec:train_details}). Evaluation is performed on the real-world datasets R1 and R2 (cf. Sec.~\ref{subsec:data}), following the evaluation framework outlined in Sec.~\ref{subsec:eval_details}. The methodology for each individual experiment is presented in detail in the following:\vspace{-0.1cm}

\paragraph{\hspace{-0.1cm}Exp. 1 -- Training Data Optimization.} To evaluate the impact of an optimized training dataset on the detection performance of YOLO-FEDER FusionNet, we build upon our previous findings presented in~\cite{Lenhard:2024}. Specifically, we adopt a training dataset that combines a large-scale, photo-realistic synthetic dataset (SynDroneVision) with a small share of real-world data, namely the publicly available dataset DUT Anti-UAV (cf. Sec.~\ref{subsec:train_details}). The resulting model is compared against the baseline YOLO-FEDER FusionNet, originally trained on the simpler synthetic dataset S1 (cf. Tab.~\ref{tab:data}), as described in \cite{Lenhard:2024_YOLO-FEDER}. For further comparison, we also include a generic YOLOv5 architecture trained using the same hybrid dataset configuration.\vspace{-0.1cm}

\begin{table}[t!]
\centering
\footnotesize
\caption{\label{tab:Scenarios_FEDER_Features}Overview of feature fusion configurations. Each configuration specifies the FEDER features integrated at the attention fusion modules (AFM 1-4, cf. Fig.~\ref{fig:yolofeder}), utilizing the final segmentation output ($\textsc{O}_S$), as well as the intermediate feature maps $\textsc{F}^D$ and $\textsc{F}^S$.}

  \begin{tabular}{clcccc}
  & & \multicolumn{4}{c}{Fusion Modules}\\\noalign{\smallskip}\cline{3-6}\noalign{\smallskip}
  & & AFM 1 & AFM 2 & AFM 3 & AFM 4\\\noalign{\smallskip}\hline\noalign{\smallskip}
 
  \multirow{10}{*}{\rotatebox{90}{\centering Feature Integration}} & Config. 1 & $\textsc{O}_S$ & $\textsc{O}_S$ & $\textsc{O}_S$ & $\textsc{O}_S$\\\noalign{\smallskip}\cline{2-6}\noalign{\smallskip}
  & Config. 2  & $\textsc{F}^S_1$ & $\textsc{F}^S_1$ & $\textsc{F}^S_1$ & $\textsc{F}^S_1$\\\noalign{\smallskip}\cline{2-6}\noalign{\smallskip}
  & Config. 3  & -- & $\textsc{F}^S_2$ & $\textsc{F}^S_3$ & $\textsc{F}^S_4$ \\\noalign{\smallskip}\cline{2-6}\noalign{\smallskip}
  & Config. 4  & -- & $\textsc{F}^D_2$ & $\textsc{F}^D_3$ & $\textsc{F}^D_4$\\\noalign{\smallskip}\cline{2-6}\noalign{\smallskip}
  & Config.  5  & -- & $\textsc{F}^S_2$, $\textsc{F}^D_2$ & $\textsc{F}^S_3$, $\textsc{F}^D_3$ & $\textsc{F}^S_4$, $\textsc{F}^D_4$\\\noalign{\smallskip}\cline{2-6}\noalign{\smallskip}
  & Config.  6  & $\textsc{O}_S$ & $\textsc{O}_S$, $\textsc{F}^D_2$ & $\textsc{O}_S$, $\textsc{F}^D_3$ & $\textsc{O}_S$, $\textsc{F}^D_4$\\\noalign{\smallskip}\hline\noalign{\smallskip}
  \end{tabular}
\end{table}

\paragraph{\hspace{-0.1cm}Exp. 2 -- Impact Analysis Intermediate FEDER Features.} By leveraging only FEDER's final output $\mathbf{O}_S$ (cf. Fig.~\ref{fig:feder}), the original YOLO-FEDER FusionNet architecture overlooks potentially valuable intermediate features generated by internal components of the FEDER branch (e.g., the DWD module or the SED, cf. Fig.~\ref{fig:feder}). To evaluate the contribution of intermediate FEDER feature representations, we extract feature maps from selected layers of the DWD module and the SED. These modules are designed to generate semantically rich hierarchical features, comparable to those obtained by the YOLO backbone. More specifically, we consider the feature maps $\mathbf{O}_S$, $\mathbf{F}^D_i$ for $i\in\{2, \ldots, 4\}$ (from the DWD module), and $\mathbf{F}^S_j$ for $j\in\{1, \ldots, 4\}$ (from the SED) across six distinct fusion scenarios (see Tab.~\ref{tab:Scenarios_FEDER_Features}). Feature fusion is performed at spatially aligned stages within the neck of YOLO-FEDER FusionNet, leveraging the hierarchical structure of FEDER features. By ensuring spatial compatibility, this alignment eliminates the need for resizing operations. This mitigates interpolation artifacts and preserves feature integrity. The integration process is facilitated by four attention-based fusion modules (AFM 1-4, Fig.~\ref{fig:yolofeder}), each implementing a two-stage mechanism consisting of feature concatenation followed by channel-wise attention. The adopted fusion strategy remains consistent with the approach originally proposed in YOLO-FEDER FusionNet \cite{Lenhard:2024_YOLO-FEDER}. An overview of the distinct fusion configurations and their associated FEDER features is provided in Tab.~\ref{tab:Scenarios_FEDER_Features}.\vspace{-0.1cm}

\paragraph{\hspace{-0.1cm}Exp. 3 -- Evaluation of YOLO-based Backbone Variations.} YOLO-FEDER FusionNet's initial iteration employs YOLOv5l as backbone for generic, multi-scale feature extraction (cf.~\cite{Lenhard:2024_YOLO-FEDER}). While this configuration has proven effective, more recent YOLO iterations (e.g., YOLOv8) have demonstrated enhanced detection accuracy on established benchmark datasets. To evaluate the architectural adaptability of YOLO-FEDER FusionNet and explore the potential benefits of incorporating more advanced feature extractors, we systematically replace the original YOLOv5l backbone with successor YOLO variants. To ensure methodological consistency, we select YOLO backbone architectures with trainable parameter counts closely aligned with that of YOLOv5l (cf. Tab.~\ref{tab:yolobckbnspecs}). Specifically, we include the backbone components of YOLOv8 (m and l), YOLOv9 (c and e), and YOLOv11 (l and x). The overall network architecture (cf. Fig.~\ref{fig:yolofeder}), the fusion strategy -- specifically Config. 4 (cf. Tab.~\ref{tab:Scenarios_FEDER_Features}) -- and the training and evaluation conditions remain unchanged.\vspace{-0.1cm}

\paragraph{\hspace{-0.1cm}Exp. 4 -- Detection Error Assessment.} Building upon the trained models and insights from Exp. 3, this experiment provides a systematic evaluation of detection errors -- focusing specifically on false positives and false negatives across distinct camera field-of-views (FOVs). The analysis combines quantitative statistical indicators -- specifically, distribution patterns and average dimensions -- with qualitative visual inspection to provide a comprehensive understanding of the network’s detection behavior. Particular emphasis is placed on the architectural design of YOLO-FEDER FusionNet,  characterized by a fixed input resolution of 640$\times$640 pixels, a hierarchical downsampling technique, and a multi-scale detection head (cf. Sec.~\ref{subsec:head}). The interplay of these architectural components introduces scale sensitivity, which can adversely affect the detection of small objects -- particularly in wide FOV configurations. To further quantify the impact of this architectural constraint, we perform an additional analysis to investigate how the exclusion of small-scale objects -- both from ground truth annotations and model predictions -- affects the overall detection performance. Following the categorization in~\cite{Hao:2025}, a minimum object size threshold of 16$\times$16 pixels is defined with respect to the original image resolution. Instances below this threshold are excluded from evaluation.

\section{\label{sec:results}Results}
A comprehensive analysis of the experimental results (cf. Sec.~\ref{sec:expsetup}) is provided in the following sections.

\begin{table}[t!]
\centering
\footnotesize
\caption{\label{tab:Train_data_var_BBW0.05}Performance comparison between YOLOv5l and YOLO-FEDER FusionNet, trained on S1 and a combination of SynDroneVision (SDV)~\cite{Lenhard:2024} and DUT Anti-UAV (DUT)~\cite{Zhao:2022}. Best results across  are highlighted in \textbf{bold}, while the second-best are \underline{underlined}.}
  \begin{tabular}{cccccccccccccc}
  & & & & &\multicolumn{4}{c}{Dataset R1} & & \multicolumn{4}{c}{Dataset R2}\\
  \noalign{\smallskip}\hline\noalign{\smallskip}
    Model & \multicolumn{2}{c}{Training Data} & Img Size & &\multicolumn{2}{c}{mAP~$\uparrow$} & FNR~$\downarrow$ & FDR~$\downarrow$ & & \multicolumn{2}{c}{mAP~$\uparrow$} & FNR~$\downarrow$ & FDR~$\downarrow$\\
    & S1 & SDV + DUT & & & @0.25 & @0.5 & & & &@0.25 & @0.5\\\noalign{\smallskip}\hline\noalign{\smallskip}
    \multirow{8}*{YOLOv5l} & \ding{51} & -- & \multirow{2}*{2040$\times$1086} && 0.572 & 0.551 & 0.463 & 0.500 & & 0.571 & 0.432 & 0.745 & 0.290\\
    & -- & \ding{51} &  & & 0.847 & 0.826 & 0.241 & \textbf{0.025} & & \underline{0.829} & 0.692 & 0.376 & 0.025
    \\\noalign{\smallskip}\cline{2-14}\noalign{\smallskip}
    
    & \ding{51} & -- & \multirow{2}*{1080$\times$1080} && 0.568 & 0.548 & 0.457 & 0.261 & & 0.685 & 0.270 & 0.473 & 0.029\\
    & -- & \ding{51} & & & \underline{0.861} & 0.819 & \underline{0.199} & 0.073 & & 0.805 & 0.645 & 0.440 & 0.038
    \\\noalign{\smallskip}\cline{2-14}\noalign{\smallskip}
    
    & \ding{51} & -- & \multirow{2}*{640$\times$640} && 0.433 & 0.401 & 0.601 & 0.311 & & 0.102 & 0.047 & 0.858 & 0.638 \\
    & -- & \ding{51} & & & 0.747 & 0.684 & 0.345 & \underline{0.040} & & 0.574 & 0.408 & 0.647 & 0.047
    \\\noalign{\smallskip}\hline\noalign{\smallskip}
     
    \multirow{5}*{YOLO-FEDER} & \ding{51} & -- & \multirow{2}*{1080$\times$1080} & & 0.708 & 0.636 & 0.449 & 0.066 & & 0.816 & 0.423 & 0.335 & \textbf{0.007}\\
    & -- & \ding{51} & & & 0.848 & \underline{0.832} & 0.205 & 0.166 & & \textbf{0.946} & \textbf{0.774} & \textbf{0.139} & 0.011 \\\noalign{\smallskip}\cline{2-14}\noalign{\smallskip}
    
    & \ding{51} & -- & \multirow{2}*{640$\times$640} & & 0.729 & 0.669 &0.372 & 0.114 & & 0.685 & 0.270 & 0.473 & 0.029\\
    & -- & \ding{51} & & & \textbf{0.879} & \textbf{0.852} & \textbf{0.197} & 0.045 &  & \textbf{0.946} & \underline{0.762} & \underline{0.146} & \underline{0.018} \\\noalign{\smallskip}\hline\noalign{\smallskip}
  \end{tabular}
\end{table}

\subsection{\label{subsec:enhanced_training_data}Training Data Optimization}
The comparative analysis in Tab.~\ref{tab:Train_data_var_BBW0.05} highlights the clear benefit of an optimized training dataset that combines large-scale, photo-realistic synthetic data (SynDroneVision~\cite{Lenhard:2024}) with limited real-world samples (DUT Anti-UAV~\cite{Zhao:2022}). Models trained on the optimized dataset consistently achieve superior performance compared to those trained exclusively on the synthetic dataset S1 across all evaluation metrics.

YOLO-FEDER FusionNet demonstrates the highest overall performance when trained on the combination of SynDroneVision and DUT Anti-UAV, particularly at a cropping resolution of 640$\times$640. In this configuration, the model achieves mAP values of 0.879 (R1) and 0.946 (R2) at an IoU threshold of 0.25 (cf. mAP@0.25, Tab.~\ref{tab:Train_data_var_BBW0.05}), along with notably low FNRs of 0.197 and 0.146, respectively. Most configurations also exhibit a decline in FDRs. However, an exception is observed on Dataset R2 at a cropping resolution of 1080$\times$1080, where YOLO-FEDER FusionNet yields a comparatively higher FDR.

In comparison to the generic YOLOv5l detection model trained under identical conditions, YOLO-FEDER FusionNet consistently yields improved performance across key performance indicators. While YOLOv5l benefits from the enhanced training data, YOLO-FEDER FusionNet matches or exceeds this performance. Simultaneously, YOLO-FEDER FusionNet, trained on the optimized dataset, offers substantially lower FNRs and FDRs -- most notably on the more challenging Dataset R2.

\subsection{\label{subsec:FEDER_Features}Impact of Intermediate FEDER Feature Representations}
A systematic evaluation of the integration of diverse FEDER feature representations (cf. Exp.~2, Sec.~\ref{subsec:exp_details}) across multiple datasets and cropping resolutions reveals notable variations in performance (see Tab.~\ref{tab:Eff_FEDERFeatures}). These differences appear to be primarily driven by the characteristics of the extracted features and the stage at which they are integrated into the network.

Among all evaluated fusion configurations, Config. 4 -- which leverages intermediate DWD features $\mathbf{F}^D_i$, $i\in\{2,\dots,4\}$ -- demonstrates the most favorable performance across all evaluation metrics. At a cropping resolution of 1080$\times$1080, it achieves the highest mAP values at both IoU thresholds (0.25 and 0.5) and obtains the highest overall fitness scores across both datasets. Despite a marginal drop in fitness at 640$\times$640 compared to the best-performing configuration (e.g., by 0.044 on Dataset R2), it remains highly competitive -- ranking second across most key metrics. With an average score of 0.882, Config. 4 achieves the highest overall fitness across all dataset-resolution combinations, indicating a strong overall trade-off between detection precision and generalization capability.

In comparison, Config. 1 -- which exclusively incorporates the final FEDER output $\mathbf{O}_S$ at each AFM block (cf. Tab.~\ref{tab:Scenarios_FEDER_Features}) -- exhibits competitive performance, particularly at the lower cropping resolution of 640$\times$640. At this resolution, it yields the highest scores across all evaluation metrics on Dataset R1. Furthermore, it ranks second in fitness when averaged  across all datasets and cropping resolutions, with an overall score of 0.873. The integration of intermediate SED features $\mathbf{F}^S_j$, $j\in\{1,\dots,4\}$ (Configs. 2 and 3) demonstrates strong detection performance on Dataset R2, with low FDRs (0.011 and 0.009) and competitive mAP values. However, both configurations exhibit notably higher FDRs on Dataset R1, with values reaching up to 0.221 (cf. Tab.~\ref{tab:Eff_FEDERFeatures}). Hence,  neither configuration surpasses the average fitness score achieved by the integration of DWD features (Config. 4).

The joint integration of DWD and SED features (Config. 5) does not yield significant performance improvements over the exclusive use of DWD-derived features. Despite achieving the lowest FNR and FDR on Dataset R2 at a cropping resolution of 640$\times$640, these improvements do not translate into enhanced overall performance. The limited effectiveness of incorporating additional SED features is also reflected in the lower average fitness score of 0.861. A similar observation can be made for Config. 6, which incorporates the final binary segmentation output $\mathbf{O}_S$ alongside DWD features. Among all fusion configurations, Config. 6 demonstrates the weakest overall performance, with an average fitness score of 0.845.

Overall, these findings indicate that leveraging intermediate DWD features (according to Config. 4) offers the most substantial contribution to enhancing the performance of YOLO-FEDER FusionNet, especially under complex environmental conditions. Accordingly, this configuration is adopted as the default feature fusion strategy for all subsequent experiments (see Secs.~\ref{subsec:results_bckbns} and \ref{subsec:results_errors}).

\begin{table}[t!]
\centering
\footnotesize
\caption{\label{tab:Eff_FEDERFeatures}Evaluation of FEDER backbone features and fusion configurations in YOLO-FEDER FusionNet (cf. Tab.~\ref{tab:Scenarios_FEDER_Features}), using a YOLOv5l backbone  trained on SynDroneVision (SDV)~\cite{Lenhard:2024} and DUT Anti-UAV (DUT)~\cite{Zhao:2022}. Best results are in \textbf{bold}, second-best are \underline{underlined}. Fitness scores are given as absolute values and relative differences (percentage points) to the baseline (Config. 1).}
  \begin{tabular}{cccccccccccccc}
  & & & \multicolumn{5}{c}{Dataset R1} & & \multicolumn{5}{c}{Dataset R2}\\
  \noalign{\smallskip}\hline\noalign{\smallskip}
    Img Size & Feature Int. &  &\multicolumn{2}{c}{mAP~$\uparrow$} & FNR~$\downarrow$ & FDR~$\downarrow$ & fitness~$\uparrow$ & & \multicolumn{2}{c}{mAP~$\uparrow$} & FNR~$\downarrow$ & FDR~$\downarrow$ & fitness~$\uparrow$ \\
    & Config. & &  @0.25 & @0.5 & & & & &@0.25 & @0.5\\\noalign{\smallskip}\hline\noalign{\smallskip}
    \multirow{6}*{1080$\times$1080} 
    & 1 & & \underline{0.848} & \underline{0.832} & \textbf{0.205} & \underline{0.166} & \multicolumn{1}{l}{\underline{0.818}} 
    & & 0.946 & 0.774 & \underline{0.139} & \underline{0.011} & \multicolumn{1}{l}{\underline{0.906}}\\
    & 2 & & 0.837 & 0.819 & \underline{0.210} & 0.221 & \multicolumn{1}{l}{0.794 \ \color{red}{($-$2.4)}}
    & &  \underline{0.953} & \underline{0.790} & 0.143 & \underline{0.011} & \multicolumn{1}{l}{\underline{0.906} \ \color{Green}{($+$0.0)}} \\
    & 3 & & 0.839 & 0.821 & 0.228 & 0.179 & \multicolumn{1}{l}{0.801 \ {\footnotesize\color{red}{($-$1.7)}}}
    & & 0.946 & 0.788 & 0.148 & \textbf{0.009} & \multicolumn{1}{l}{0.904 \ \color{red}{($-$0.2)}} \\
    & 4 & & \textbf{0.853} & \textbf{0.835} & 0.220  & \textbf{0.094} & \multicolumn{1}{l}{\textbf{0.837} \ \color{Green}{($+$1.9)}}
    & &  \textbf{0.962} & \textbf{0.828} & \textbf{0.092} & 0.013 & \multicolumn{1}{l}{\textbf{0.933} \ \color{Green}{($+$2.7)}} \\
    & 5 & & 0.816 & 0.800 & 0.225 & 0.294 & \multicolumn{1}{l}{0.758 \ {\footnotesize\color{red}{($-$6.0)}}}
    & & 0.937 & 0.755 & 0.155 & 0.012 & \multicolumn{1}{l}{0.895 \ {\footnotesize\color{Red}{($-$1.1)}}} \\
    & 6 & & 0.834 & 0.819 & 0.224 & 0.219 &  \multicolumn{1}{l}{0.788 \ \color{Red}{($-$3.0)}}
    & & 0.935 & 0.750 & 0.162 & 0.015 & \multicolumn{1}{l}{0.890 \ \color{Red}{($-$1.6)}} \\\noalign{\smallskip}\hline\noalign{\smallskip}
    \multirow{6}*{640$\times$640} 
    & 1 & & \textbf{0.879} & \textbf{0.852} & \textbf{0.197} & \textbf{0.045} & \multicolumn{1}{l}{\textbf{0.869}} 
    & & 0.946 & 0.762 & 0.146 & 0.018 & \multicolumn{1}{l}{0.899}\\
    & 2 & & \underline{0.858} & \underline{0.832} & \underline{0.213} & 0.112 & \multicolumn{1}{l}{0.834 \ \color{Red}{($-$3.5)}} 
    & & \textbf{0.953} & \underline{0.798} & 0.156 & \underline{0.007} & \multicolumn{1}{l}{0.903 \ \color{Green}{($+$0.4)}} \\
    & 3 & & 0.838 & 0.813 & 0.244 & 0.068 & \multicolumn{1}{l}{0.832 \ \color{Red}{($-$3.7)}} 
    & & \underline{0.948} & 0.788 & 0.154 & 0.009 & \multicolumn{1}{l}{0.901 \ \color{Green}{($+$0.5)}} \\
    & 4 & & 0.853 & 0.830 & 0.229 & \underline{0.055} & \multicolumn{1}{l}{\underline{0.846} \ \color{Red}{($-$2.3)}}
    & & 0.944 & \textbf{0.804} & \underline{0.121} & 0.026 & \multicolumn{1}{l}{\underline{0.911} \ \color{Green}{($+$1.2)}} \\
    & 5 & & 0.851 & 0.830 & 0.233 & 0.077 & \multicolumn{1}{l}{0.836 \ \color{Red}{($-$3.3)}}
    & & 0.921 & 0.759 & \textbf{0.025} & \textbf{0.005} & \multicolumn{1}{l}{\textbf{0.955} \ \color{Green}{($+$5.6)}} \\
    & 6 & & 0.853 & \underline{0.832} & 0.220 & 0.089 & \multicolumn{1}{l}{0.838 \ \color{Red}{($-$3.1)}}
    & & 0.912 & 0.730 & 0.166 & 0.070 & \multicolumn{1}{l}{0.865 \ \color{Red}{($-$3.4)}} \\\noalign{\smallskip}\hline\noalign{\smallskip}
  \end{tabular}
\end{table}

\begin{table}[t!]
\centering
\footnotesize
\caption{\label{tab:resultsYOLOFEDER_bckbns}Performance evaluation of different YOLO backbones within the YOLO-FEDER FusionNet framework. The evaluated architectures follow the neck and head design depicted in Fig.~\ref{fig:yolofeder}, leveraging FEDER features $F^D_i$, $i\in\{2,\dots,4\}$ according to feature fusion configuration 4 (cf. Tab.~\ref{tab:Scenarios_FEDER_Features}). Best results are in \textbf{bold}, second-best are \underline{underlined}. Fitness scores are given as absolute values and relative differences (percentage points) to the baseline (YOLO backbone v5l).}
  \begin{tabular}{clccccclcccccl}
  & & & \multicolumn{5}{c}{Dataset R1} & & \multicolumn{5}{c}{Dataset R2}\\
  \noalign{\smallskip}\hline\noalign{\smallskip}
    Img Size & \multicolumn{1}{c}{YOLO} & & \multicolumn{2}{c}{mAP~$\uparrow$} & FNR~$\downarrow$ & FDR~$\downarrow$ & fitness~$\uparrow$ & & \multicolumn{2}{c}{mAP~$\uparrow$} & FNR~$\downarrow$ & FDR~$\downarrow$ & fitness~$\uparrow$ \\
    & \multicolumn{1}{c}{Backbone} & & @0.25 & @0.5 & & & & & @0.25 & @0.5\\\noalign{\smallskip}\hline\noalign{\smallskip}
    
    \multirow{7}*{1080$\times$1080} & \hspace{0.4cm}v5l & & \underline{0.853} & 0.835 & 0.220 & 0.094 & \multicolumn{1}{l}{0.837} 
    & & \underline{0.962} & \textbf{0.828} & \textbf{0.092} & 0.013 & \multicolumn{1}{l}{\textbf{0.933}} \\
    
    & \hspace{0.4cm}v8m & & 0.838 & 0.824 & \underline{0.208} & 0.258 & \multicolumn{1}{l}{0.783 \ \color{Red}{($-$5.4)}} 
    & & 0.959 & \underline{0.817} & 0.112 & \textbf{0.008} & \multicolumn{1}{l}{0.924 \ \color{Red}{($-$0.9)}}\\
    
    & \hspace{0.4cm}v8l & & \textbf{0.872} & \textbf{0.852} & \textbf{0.192} & 0.049 & \multicolumn{1}{l}{\textbf{0.869} \ \color{Green}{($+$3.2)}} 
    & & \textbf{0.967} & 0.810 & \underline{0.107} & \underline{0.012} & \multicolumn{1}{l}{0.925 \ \color{Red}{($-$0.8)}}\\
    
    & \hspace{0.4cm}v9c & & 0.815 & 0.795 & 0.235 & 0.184 & \multicolumn{1}{l}{0.791 \ \color{Red}{($-$4.6)}} 
    & & 0.953 & 0.802 & \textbf{0.092} & 0.020 & \multicolumn{1}{l}{\underline{0.927} \ \color{Red}{($-$0.6)}}\\
    
    & \hspace{0.4cm}v9e & & 0.852 & \underline{0.838} & 0.215 & 0.086 & \multicolumn{1}{l}{\underline{0.842} \ \color{Green}{($+$0.5)}} 
    & & 0.940 & 0.800 & 0.109 & 0.024 & \multicolumn{1}{l}{0.917 \ \color{Red}{($-$1.6)}}\\
    
    & \hspace{0.4cm}v11l & & 0.828 & 0.808 & 0.293 & \underline{0.029} & \multicolumn{1}{l}{0.822 \ \color{Red}{($-$1.5)}} 
    & & 0.896 & 0.760 & 0.138 & 0.163 & \multicolumn{1}{l}{0.847 \ \color{Red}{($-$8.6)}}\\
    
    & \hspace{0.4cm}v11x & & 0.803 & 0.777 & 0.333 & \textbf{0.013} & \multicolumn{1}{l}{0.804 \ \color{Red}{($-$3.3)}} 
    & & 0.938 & 0.784 & 0.110 & 0.071 & \multicolumn{1}{l}{0.898 \ \color{Red}{($-$3.5)}}\\\noalign{\smallskip}\hline\noalign{\smallskip}
    
    \multirow{7}*{640$\times$640} & \hspace{0.4cm}v5l & & 0.853 & 0.830 & 0.229 & \underline{0.055} & \multicolumn{1}{l}{\underline{0.846}} 
    & & 0.944 & \underline{0.804} & 0.121 & 0.026 & \multicolumn{1}{l}{0.911} \\
    
    & \hspace{0.4cm}v8m  & & 0.836 & 0.813 & 0.211 & 0.173 & \multicolumn{1}{l}{0.809 \ \color{Red}{($-$3.7)}} 
    & & 0.949 & 0.710 & 0.113 & 0.039 & \multicolumn{1}{l}{0.901 \ \color{Red}{($-$1.0)}}\\
    
    & \hspace{0.4cm}v8l  & & \textbf{0.876} & \textbf{0.848} & \underline{0.188} & 0.116 & \multicolumn{1}{l}{\textbf{0.847} \ \color{Green}{($+$0.1)}} 
    & & \textbf{0.967} & 0.798 & \textbf{0.082} & \underline{0.019} & \multicolumn{1}{l}{\textbf{0.933} \ \color{Green}{($+$2.2)}}\\
    
    & \hspace{0.4cm}v9c  & & 0.830 & 0.806 & 0.242 & 0.095 & \multicolumn{1}{l}{0.822 \ \color{Red}{($-$2.4)}} 
    & & \underline{0.959} & \textbf{0.806} & 0.111 & \textbf{0.007} & \multicolumn{1}{l}{\underline{0.924} \ \color{Green}{($+$1.3)}}\\
    
    & \hspace{0.4cm}v9e  & & \underline{0.863} & \underline{0.842} & \textbf{0.179} & 0.173 & \multicolumn{1}{l}{0.829 \ \color{Red}{($-$1.7)}} 
    & & 0.943 & 0.770 & \underline{0.105} & 0.040 & \multicolumn{1}{l}{0.910 \ \color{Red}{($-$0.1)}}\\
    
    & \hspace{0.4cm}v11l & & 0.847 & 0.822 & 0.250 & \textbf{0.040} & \multicolumn{1}{l}{0.840 \ \color{Red}{($-$0.6)}} 
    & & 0.883 & 0.723 & 0.166 & 0.097 & \multicolumn{1}{l}{0.852 \ \color{Red}{($-$5.9)}}\\
    
    & \hspace{0.4cm}v11x & & 0.819 & 0.794 & 0.268 & 0.064 & \multicolumn{1}{l}{0.818 \ \color{Red}{($-$2.8)}} & & 0.935 & 0.754 & 0.102 & 0.078 & \multicolumn{1}{l}{0.896 \ \color{Red}{($-$1.5)}}\\\hline
  \end{tabular}
\end{table}

\subsection{\label{subsec:results_bckbns}Effectiveness of YOLO-based Backbone Variations}
Among all YOLO backbone configurations evaluated within the YOLO-FEDER FusionNet architecture, the incorporation of YOLOv8l demonstrates the highest and most consistent detection performance across all experimental conditions (cf. Tab.~\ref{tab:resultsYOLOFEDER_bckbns}). The YOLOv8l-enhanced YOLO-FEDER FusionNet achieves consistently high mAP values at an IoU threshold of 0.25 across all cropping resolutions and datasets. It also leads in mAP at an IoU threshold of 0.5 on Dataset R1. In addition, it exhibits strong overall fitness -- with an average score of 0.894 -- while maintaining low FNRs and FDRs. Specifically, it ranks among the top two configurations in terms FNR, with values ranging from 0.082 (R2, 640$\times$640) to 0.192 (R1, 1080$\times$1080). 
Despite the clear performance advantages provided by the YOLOv8l backbone, the original YOLO-FEDER FusionNet configuration -- featuring YOLOv5l -- remains a strong and reliable baseline. It achieves the second-highest average fitness score (0.882) and continues to perform competitively across other performance indicators. Nevertheless, it exhibits slightly elevated FNRs and FDRs than the YOLOv8l-based variant across the majority of evaluation scenarios. Additionally, it entails a higher model complexity, with a parameter count of 26.6M (cf. Tab.~\ref{tab:yolobckbnspecs}). 

Similarly, the YOLOv9e-based configuration of YOLO-FEDER FusionNet achieves competitive results, particularly in terms of mAP and FNR, with the lowest observed FNR on Dataset R1. However, the strong localized performance fails to generalize into comprehensive performance improvements across all indicators and datasets (average fitness score: 0.875). Furthermore, the relatively high model complexity of the YOLOv9e backbone (30.2M parameters, 117.2B GFLOPs) adversely affects the computational efficiency of YOLO-FEDER FusionNet -- especially in comparison to more balanced alternatives (e.g., YOLOv8l). 

While the YOLOv8m-based configuration of YOLO-FEDER FusionNet represents the most efficient variant in terms of parameter count (11.8M) and computational cost (38.8B GFLOPs), it exhibits a noticeable decline in performance compared to larger-backbone configurations. This degradation is particularly pronounced in terms of FDR, reaching values of 0.173 and 0.258, respectively (cf. Dataset R1, Tab.~\ref{tab:resultsYOLOFEDER_bckbns}). The lower average fitness score of 0.855 further reflects this underperformance,  ranking it among the bottom three configurations. A comparable trend is observed for YOLO-FEDER FusionNet configured with a YOLOv9c backbone (cf. Tab.~\ref{tab:resultsYOLOFEDER_bckbns}). The integration of YOLOv11l and YOLOv11x backbones into YOLO-FEDER FusionNet yields the weakest performance across most evaluation metrics, with average fitness scores of 0.840 and 0.854, respectively.

\subsection{\label{subsec:results_errors}Detection Error Assessment}
\begin{figure*}[t!]
\centering
  \includegraphics[width=0.99\textwidth, trim={0.3cm 22.8cm 0.3cm 0cm}, clip]{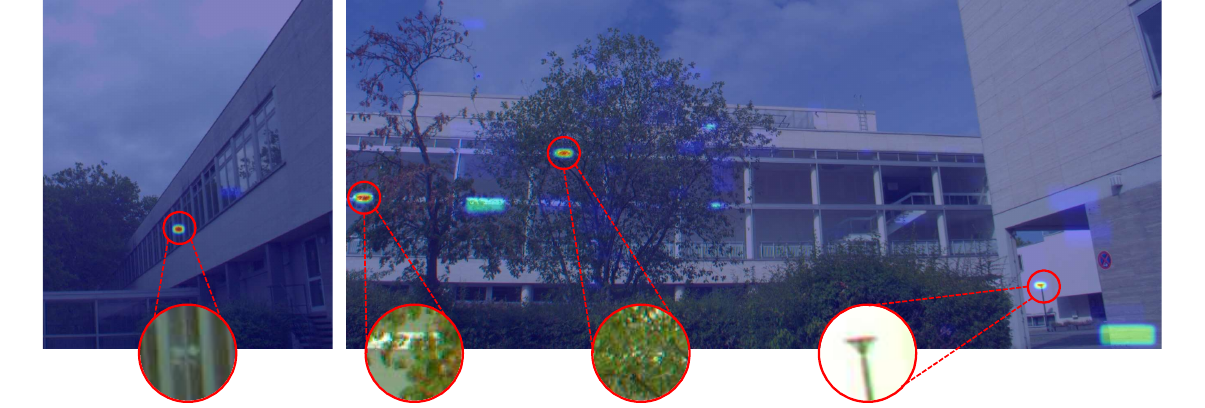}
  \caption{\label{fig:FPs_Heatmap}Spatial distribution of false positives across the two camera FOVs in Dataset R1. (Images depict cropped regions of the full FOVs.) Zoomed-in areas highlight regions with high concentrations of recurring false positives across the majority of test samples. Color intensity denotes false detection frequency, from low (blue) to high (red). Results are based on YOLO-FEDER FusionNet with a YOLOv8m backbone.}
\end{figure*}

\begin{figure}[t!]
\centering
  \includegraphics[width=0.48\textwidth, trim={0.3cm 20.5cm 10cm 0cm}, clip]{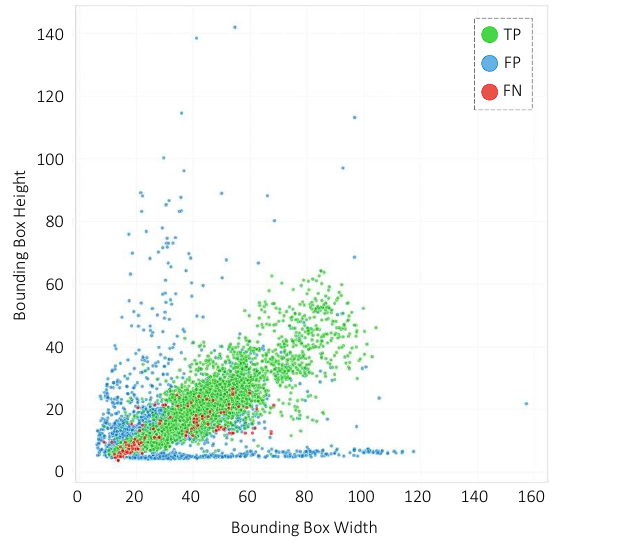}
  \caption{\label{fig:FPs_Scatter}Bounding box width vs. height for YOLO-FEDER FusionNet (YOLOv8m backbone) on R1 (1080$\times$1080). FPs seem to be well-separable from TPs based on their smaller spatial dimensions. FNs exhibit substantially overlap with TPs, indicating that FNs arise from factors beyond object scale.}
\end{figure}

As previously noted (Sec.~\ref{subsec:results_bckbns}), YOLO-FEDER FusionNet configurations with YOLOv8m and YOLOv9c backbones exhibit elevated FDRs, with the highest rates observed on Dataset R1 (cf. Tab.~\ref{tab:resultsYOLOFEDER_bckbns}). A visual analysis of the FPs, conducted separately for each camera FOV, reveals systematic clustering of FPs within specific image regions (see Fig.~\ref{fig:FPs_Heatmap}). These regions primarily align with image areas affected by visual artifacts, including reflections, irregular texture patterns, or fine structural background details, which can create drone-like signatures. Further analysis of the bounding box dimensions indicates that FPs primarily correspond to small-scale artifacts, characterized by reduced width and height (see Fig.~\ref{fig:FPs_Scatter}). For instance, the YOLOv8m-based YOLO-FEDER FusionNet produces FP bounding boxes with average dimensions of 21.2 pixels in width and 10.97 pixels in height. In contrast, true positive detections exhibit mean width and height values of 44.28 and 21.51 pixels. A similar pattern is observed for the YOLOv9c-based backbone configuration.

Motivated by these observations, we introduce a size-based filtering strategy to mitigate erroneous detections, particularly those caused by small-scale artifacts. Applying a minimum object size threshold during post-processing -- on both predictions and ground truth -- leads to substantial improvements in error reduction. For instance, the FDR of the YOLOv8m-based YOLO-FEDER FusionNet drops from 0.258 to 0.067 on Dataset R1 (1080×1080 resolution; cf. Tabs.~\ref{tab:resultsYOLOFEDER_bckbns} and \ref{tab:resultsYOLOFEDER_bckbns_OBJs_smaller_10px}). Similarly, the FDR of the YOLOv9c-based variant decreases significantly from 0.184 to 0.018. Excluding small-scale detections and corresponding ground truth instances also improves detection performance in terms of mAP across all configurations. For instance, the YOLOv8l-based YOLO-FEDER FusionNet achieves mAP values of 0.914 (R1) and 0.942 (R2) at a cropping resolution of 1080$\times$1080, while simultaneously reducing FNRs to 0.159 and 0.111, respectively. Furthermore, fitness scores improve substantially, reaching values up to 0.936 (R2, 640$\times$640), surpassing previous bests (cf. Exp.~3, Tab.~\ref{tab:resultsYOLOFEDER_bckbns}).  Despite consistent improvements across all backbone configurations, the YOLOv8l-based YOLO-FEDER FusionNet maintains the highest average fitness score of 0.912.

\begin{table*}[t!]
\centering
\footnotesize
\caption{\label{tab:resultsYOLOFEDER_bckbns_OBJs_smaller_10px}Performance evaluation of YOLO-FEDER FusionNet configurations (distinguished by backbone variations), considering only objects exceeding $16 \times 16$ pixels in size. The network architecture follows the design shown in Fig.~\ref{fig:yolofeder}, integrating FEDER features according to Config. 4. Best results are highlighted in \textbf{bold}, second-best are \underline{underlined}.}
  
  \begin{tabular}{clcccccccccccc}
  & & & \multicolumn{5}{c}{Dataset R1} & & \multicolumn{5}{c}{Dataset R2}\\
  \noalign{\smallskip}\hline\noalign{\smallskip}
    Img Size & \multicolumn{1}{c}{YOLO} & & \multicolumn{2}{c}{mAP~$\uparrow$} & FNR~$\downarrow$ & FDR~$\downarrow$ & fitness~$\uparrow$ & & \multicolumn{2}{c}{mAP~$\uparrow$} & FNR~$\downarrow$ & FDR~$\downarrow$ & fitness~$\uparrow$ \\
    & \multicolumn{1}{c}{Backbone} & & @0.25 & @0.5 & & & & & @0.25 & @0.5\\\noalign{\smallskip}\hline\noalign{\smallskip}
    \multirow{7}*{1080$\times$1080} 
    & \hspace{0.4cm}v5l  & & 0.881 & 0.877 & 0.223 & 0.016 & 0.870 & & 0.927 & 0.798 & 0.139 & \underline{0.010} & 0.906\\
    & \hspace{0.4cm}v8m  & & \underline{0.896} & \underline{0.889} & \underline{0.184} & 0.067 & \underline{0.872} & & 0.940 & 0.825 & 0.118 & 0.008 & \underline{0.921}\\
    & \hspace{0.4cm}v8l  & & \textbf{0.914} & \textbf{0.905} & \textbf{0.159} & \underline{0.017} & \textbf{0.904} & & \underline{0.942} & 0.811 & \underline{0.111} & 0.012 & \underline{0.921}\\
    & \hspace{0.4cm}v9c  & & 0.875 & 0.868 & 0.235 & 0.018 & 0.862 & & \textbf{0.946} & \textbf{0.839} & \textbf{0.103} & \textbf{0.009} & \textbf{0.929}\\
    & \hspace{0.4cm}v9e  & & 0.890 & 0.887 & 0.186 & 0.075 & 0.868 & & 0.937 & 0.838 & \underline{0.115} & 0.024 & 0.917\\
    & \hspace{0.4cm}v11l & & 0.876 & 0.864 & 0.220 & 0.028 & 0.865 & & 0.890 & 0.786 & 0.138 & 0.163 & 0.849\\
    & \hspace{0.4cm}v11x & & 0.852 & 0.842 & 0.272 & \textbf{0.011} & 0.843 & & 0.929 & 0.821 & 0.117 & 0.070 & 0.898
    \\\noalign{\smallskip}\hline\noalign{\smallskip}

    \multirow{7}*{640$\times$640} 
    & \hspace{0.4cm}v5l  & & 0.882 & 0.876 & 0.227 & \underline{0.009} & 0.870 & & 0.911 & 0.770 & 0.111 & 0.012 & 0.914\\
    & \hspace{0.4cm}v8m  & & \textbf{0.936} & 0.761 & \textbf{0.115} & 0.028 & \textbf{0.908} & & 0.936 & 0.761 & 0.115 & 0.029 & 0.908\\
    & \hspace{0.4cm}v8l  & & 0.896 & \underline{0.884} & 0.202 & \underline{0.009} & 0.884 & & \textbf{0.955} & \underline{0.822} & \textbf{0.085} & \underline{0.009} & \textbf{0.936}\\
    & \hspace{0.4cm}v9c  & & 0.871 & 0.863 & 0.252 & \textbf{0.005} & 0.858 & & \underline{0.942} & \textbf{0.834} & 0.113 & \textbf{0.007} & \underline{0.924}\\
    & \hspace{0.4cm}v9e  & & \underline{0.911} & \textbf{0.902} & \underline{0.164} & 0.028 & \underline{0.898} & & 0.941 & 0.810 & 0.106 & 0.039 & 0.914\\
    & \hspace{0.4cm}v11l & & 0.893 & 0.883 & 0.196 & 0.014 & 0.885 & & 0.874 & 0.760 & 0.165 & 0.090 & 0.858\\
    & \hspace{0.4cm}v11x & & 0.880 & 0.870 & 0.227 & 0.015 & 0.868 & & 0.936 & 0.809 & \underline{0.103} & 0.061 & 0.907\\\hline
  \end{tabular}
\end{table*}

\subsection{\label{subsec:discussion}Discussion}
The results presented in Secs.~\ref{subsec:enhanced_training_data}-\ref{subsec:results_errors} demonstrate that systematic optimization of both training data and network architecture substantially improves detection accuracy (cf. Tab.~\ref{tab:Total_Improvements}, Supp. Material). The combination of large-scale synthetic data with a limited amount of real-world samples (i.e., SynDroneVision~\cite{Lenhard:2024} and DUT Anti-UAV~\cite{Zhao:2022}) significantly enhances the performance of YOLO-FEDER FusionNet, underscoring the importance of dataset diversity and realism (cf. Sec.~\ref{subsec:enhanced_training_data}). These findings are consistent with prior work~\cite{Lenhard:2024}.

Systematic evaluation of FEDER feature representations shows that integrating intermediate DWD features yields the most consistent performance gains under our evaluation setup (cf. Sec.~\ref{subsec:FEDER_Features}). This highlights the importance of leveraging deeper, multi-scale feature information rather than relying solely on final outputs or auxiliary segmentation cues -- as previously done in~\cite{Lenhard:2024_YOLO-FEDER}. It also indicates that intermediate DWD features seem to offer more robust discriminative cues for detecting camouflaged drones compared to later-stage feature representations.

When considering backbone variations, the YOLOv8l-enhanced YOLO-FEDER FusionNet consistently achieves the most favorable trade-off between detection accuracy and computational efficiency (cf. Sec.~\ref{subsec:results_bckbns}). While larger backbones such as YOLOv9e offer competitive accuracy, their increased complexity can hinder deployment in resource-constrained environments. Conversely, lightweight variants like YOLOv8m and YOLOv9c incur higher FDRs, reflecting a trade-off between model efficiency and detection reliability. Despite supporting a variety of YOLO backbones, YOLO-FEDER FusionNet does not necessarily benefit from the superior benchmark performance of newer models (e.g., YOLOv11). This effect may stem from the decoupled use of backbone components, which potentially hinders the network’s ability to leverage the full representational and optimization benefits of YOLO’s end-to-end architecture -- benefits that are specifically tuned for maximizing performance on benchmark data.

A detailed analysis of FPs reveals that small-scale artifacts constitute a primary source of detection errors (cf. Sec.~\ref{subsec:results_errors}). This behavior likely arises from YOLO-FEDER FusionNet’s multi-scale detection strategy, which predicts drone instances at three spatial resolutions -- 20$\times$20, 40$\times$40, and 80$\times$80 -- for an input size of 640$\times$640. Successive downsampling limits reliable detection of objects smaller than 8$\times$8 pixels and reduces feature discriminability in regions below $\sim$16$\times$16 pixels~\cite{Lyu:2023}. This seems to lead to erroneous activations triggered by background noise and visual artifacts. Incorporating a size-filtering post-processing step mitigates this issue by reducing FDRs -- particularly in lightweight backbone configurations -- while also improving mAP and overall fitness. These findings reinforce the adverse impact of small drone instances on detection performance~\cite{Lv:2022,Liu:2021,Chen:2024,Kim:2023,Dong:2023}. Enhancing YOLO-FEDER FusionNet with an additional detection scale at 160$\times$160, following~\cite{Chen:2024}, offers a promising direction to overcome current architectural constraints.

Finally, it should be noted that the most favorable configuration of YOLO-FEDER FusionNet is inherently tied to the weighting of performance metrics, which may vary by application and may require task-specific re-evaluation.

\section{\label{sec:conclusion}Conclusion}
In this work, we presented an enhanced version of YOLO-FEDER FusionNet for robust drone detection in visually complex environments. By systematically optimizing training data, FEDER feature integration, and YOLO backbone architecture, we achieved notable gains in detection accuracy and robustness. Our findings highlight the importance of combining large-scale, photo-realistic synthetic training data with real-world samples, leveraging intermediate multi-scale FEDER features, and aligning post-processing strategies with architectural constraints. For our target application, YOLO-FEDER FusionNet configured with a YOLOv8l backbone and intermediate DWD features demonstrated the most favorable performance. Future work will explore architectural refinements, such as higher-resolution detection heads and improved feature aggregation, to boost performance in complex scenes.

\bibliographystyle{unsrt}
\bibliography{references}

% SUPPLEMENTARY MATERIAL
\newpage
\vbox{%
    \hsize\textwidth
    \linewidth\hsize
    \vskip 0.5in
    %\@toptitlebar
    \centering
    {\LARGE\sc Supplementary Material\par}
    %\@bottomtitlebar
    %\textsc{\undertitle}\\
    \vskip 0.5in
}

\appendix

\section{\label{supp:module_comparison} Comparison of C3, C2F, and C3K2 Modules}
\begin{figure*}[ht!]
\centering
  \includegraphics[width=0.99\textwidth, trim={0.3cm 22.5cm 0.3cm 0cm}, clip]{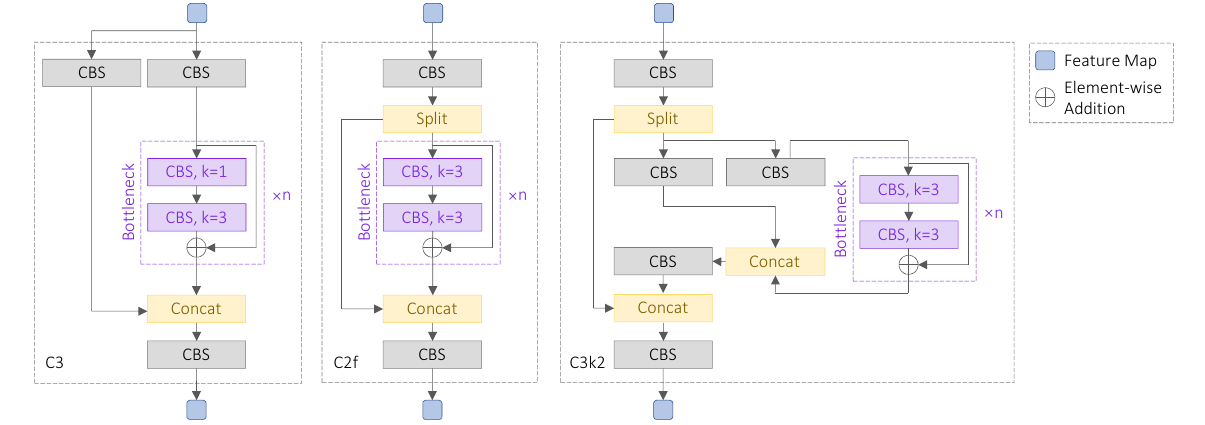}
  \caption{\label{fig:ModuleComparison} Structural comparison of the C3 (left), C2f (middle), and C3K2 (right) modules, highlighting key architectural differences in block composition and connectivity. Layer abbreviations are as follows: CBS (convolution, batch normalization, and SiLU activation), and Concat (concatenation layer). In CBS modules, $k$ denotes the kernel size of the convolutional layers, while in bottleneck structures, $n$ indicates the number of repetitions. The Split layer partitions the input feature map equally along the channel dimension; one half is forwarded directly to the Concat layer, while the other half undergoes further transformation through intermediate operations.
}
\end{figure*}

This section presents a visual comparison of the C3 (YOLOv5), C2f (YOLOv8), and C3k2 (YOLOv11) modules, as shown in Fig.~\ref{fig:ModuleComparison}, emphasizing their structural differences and architectural design characteristics. Here, $k$ represents the convolutional kernel size, and $n$ denotes the number of module repetitions, which is typically set to one.

\begin{table*}[t]
\centering\footnotesize
\caption{Performance comparison of YOLO-FEDER FusionNet trained on a combination of SynDroneVision (SDV)~\cite{Lenhard:2024} and DUT Anti-UAV~\cite{Zhao:2022} using different bounding box loss weights. Evaluation is conducted on real-world datasets R1 and R2. The best result per dataset and metric is shown in \textbf{bold}, the second-best is \underline{underlined}, and the top four are highlighted in {\color{SkyBlueCB}blue}.
}
  \label{tab:bboox_loss_weight}
  \begin{tabular}{cccccccccccccc}
  
  & & &\multicolumn{5}{c}{Dataset R1} & & \multicolumn{5}{c}{Dataset R2}\\
  \noalign{\smallskip}\hline\noalign{\smallskip}
    Img  Size & Bbox Loss Weight & &\multicolumn{2}{c}{mAP~$\uparrow$} & FNR~$\downarrow$ & FDR~$\downarrow$ & fitness~$\uparrow$ & & \multicolumn{2}{c}{mAP~$\uparrow$} & FNR~$\downarrow$ & FDR~$\downarrow$ & fitness~$\uparrow$ \\
    & $w_1$ & &  @0.25 & @0.5 & & & & &@0.25 & @0.5\\\noalign{\smallskip}\hline\noalign{\smallskip}

    \multirow{14}*{1080$\times$1080} & 7.50 & & 0.820 & 0.805 & 0.247 & 0.220 & 0.774 & & 0.931 & 0.755 & 0.183 & {\color{SkyBlueCB}\underline{0.010}} & 0.883\\
    & 6.00 & & 0.793 & 0.777 & 0.249 & 0.299 & 0.740 & & 0.922 & 0.730 & 0.186 & 0.014 & 0.877\\
    & 4.50 & & 0.834 & 0.817 & 0.242 & 0.162 & 0.800 & & 0.933 & 0.747 & 0.192 & 0.009 & 0.879\\
    & 3.00 & & 0.813 & 0.796 & 0.259 & 0.225 & 0.766 & & 0.938 & {\color{SkyBlueCB}\textbf{0.792}} & 0.143 & {\color{SkyBlueCB}\textbf{0.008}} & {\color{SkyBlueCB}0.906}\\
    & 1.50 & & {\color{SkyBlueCB}\underline{0.849}} & {\color{SkyBlueCB}0.828} & 0.226 & {\color{SkyBlueCB}\textbf{0.073}} & {\color{SkyBlueCB}\textbf{0.841}} & & 0.937 & 0.749 & 0.164 & 0.016 & 0.889\\
    & 1.25 & & 0.824 & 0.804 & 0.229 & 0.218 & 0.784 & & 0.930 & 0.746 & 0.181 & {\color{SkyBlueCB}0.011} & 0.882\\
    & 1.00 & & {\color{SkyBlueCB}0.844} & {\color{SkyBlueCB}0.826} & {\color{SkyBlueCB}\underline{0.208}} & 0.175 & 0.812 & & {\color{SkyBlueCB}0.949} & {\color{SkyBlueCB}\underline{0.788}} & {\color{SkyBlueCB}\textbf{0.121}} & 0.019 & {\color{SkyBlueCB}\underline{0.913}}\\
    & 0.75 & & 0.827 & 0.808 & 0.233 & 0.174 & 0.798 & & 0.944 & 0.760 & 0.153 & 0.016 & 0.896\\
    & 0.50 & & {\color{SkyBlueCB}\textbf{0.854}} & {\color{SkyBlueCB}\textbf{0.833}} & 0.229 & {\color{SkyBlueCB}\underline{0.077}} & {\color{SkyBlueCB}\underline{0.839}} & & 0.937 & 0.771 & 0.144 & 0.021 & 0.899\\
    & 0.25 & & 0.843 & 0.825 & 0.238 & {\color{SkyBlueCB}0.078} & {\color{SkyBlueCB}0.832} & & 0.941 & 0.770 & 0.168 & {\color{SkyBlueCB}0.011} & 0.892\\
    & 0.20 & & 0.839 & 0.821 & {\color{SkyBlueCB}0.216} & 0.147 & 0.817 & & {\color{SkyBlueCB}\textbf{0.953}} & {\color{SkyBlueCB}0.778} & 0.134 & 0.012 & {\color{SkyBlueCB}0.909}\\
    & 0.15 & & 0.839 & 0.822 & {\color{SkyBlueCB}0.217} & 0.206 & 0.796 & & 0.939 & 0.763 & {\color{SkyBlueCB}0.142} & 0.013 & 0.902\\
    & 0.10 & & {\color{SkyBlueCB}0.848} & {\color{SkyBlueCB}\underline{0.832}} & {\color{SkyBlueCB}\textbf{0.205}} & 0.166 & 0.818 & & {\color{SkyBlueCB}0.946} & 0.774 & {\color{SkyBlueCB}0.139} & {\color{SkyBlueCB}0.011} & {\color{SkyBlueCB}0.906}\\
    & 0.05 & & 0.840 & 0.821 & 0.237 & {\color{SkyBlueCB}0.103} & {\color{SkyBlueCB}0.823} & & {\color{SkyBlueCB}\underline{0.952}} & {\color{SkyBlueCB}0.784} & {\color{SkyBlueCB}\underline{0.122}} & 0.012 & {\color{SkyBlueCB}\textbf{0.915}} \\\noalign{\smallskip}\hline\noalign{\smallskip}

    \multirow{14}*{640$\times$640} & 7.50 & & 0.848 & 0.822 & 0.234 & 0.082 & 0.833 & & {\color{SkyBlueCB}0.945} & 0.762 & 0.141 & {\color{SkyBlueCB}\underline{0.013}} & 0.903\\
    & 6.00 & & 0.821 & 0.789 & 0.256 & 0.111 & 0.807 & & 0.908 & 0.722 & 0.212 & 0.032 & 0.856\\
    & 4.50 & & 0.852 & 0.827 & 0.232 & 0.077 & 0.837 & & 0.938 & 0.764 & 0.176 & 0.020 & 0.884\\
    & 3.00 & & 0.833 & 0.803 & 0.258 & 0.071 & 0.823 & & 0.939 & {\color{SkyBlueCB}\underline{0.789}} & 0.143 & {\color{SkyBlueCB}0.019} & {\color{SkyBlueCB}0.902}\\
    & 1.50 & & 0.864 & {\color{SkyBlueCB}0.836} & 0.215 & \textbf{0.056} & {\color{SkyBlueCB}\underline{0.854}} & & 0.931 & 0.742 & 0.167 & 0.027 & 0.883\\
    & 1.25 & & {\color{SkyBlueCB}\underline{0.871}} & {\color{SkyBlueCB}\underline{0.845}} & {\color{SkyBlueCB}0.213} & {\color{SkyBlueCB}0.067} & {\color{SkyBlueCB}0.852} & & 0.942 & 0.761 & 0.147 & 0.025 & 0.895\\
    & 1.00 & & 0.861 & 0.834 & {\color{SkyBlueCB}0.211} & 0.097 & 0.841 & & {\color{SkyBlueCB}\textbf{0.950}} & 0.776 & {\color{SkyBlueCB}\underline{0.127}} & 0.024 & {\color{SkyBlueCB}\textbf{0.907}}\\
    & 0.75 & & 0.852 & 0.826 & 0.235 & 0.086 & 0.832 & & 0.942 & {\color{SkyBlueCB}0.787} & {\color{SkyBlueCB}0.139} & 0.037 & 0.897\\
    & 0.50 & & {\color{SkyBlueCB}0.867} & {\color{SkyBlueCB}0.843} & 0.219 & {\color{SkyBlueCB}0.064} & 0.850 & & 0.944 & {\color{SkyBlueCB}0.785} & {\color{SkyBlueCB}\textbf{0.121}} & 0.037 & {\color{SkyBlueCB}\underline{0.906}}\\
    & 0.25 & & 0.855 & 0.832 & 0.235 & {\color{SkyBlueCB}\underline{0.051}} & 0.845 & & 0.938 & {\color{SkyBlueCB}\underline{0.789}} & 0.205 & {\color{SkyBlueCB}\textbf{0.012}} & 0.876\\
    & 0.20 & & {\color{SkyBlueCB}0.867} & 0.838 & {\color{SkyBlueCB}\underline{0.204}} & 0.076 & {\color{SkyBlueCB}0.852} & & 0.944 & 0.777 & 0.147 & {\color{SkyBlueCB}0.019} & 0.899\\
    & 0.15 & & 0.843 & 0.821 & 0.234 & 0.126 & 0.817 & & 0.939 & {\color{SkyBlueCB}\textbf{0.791}} & 0.164 & 0.020 & 0.892\\
    & 0.10 & & {\color{SkyBlueCB}\textbf{0.879}} & {\color{SkyBlueCB}\textbf{0.852}} & {\color{SkyBlueCB}\textbf{0.197}} & {\color{SkyBlueCB}\textbf{0.045}} & {\color{SkyBlueCB}\textbf{0.869}} & & {\color{SkyBlueCB}0.946} & 0.762 & 0.146 & {\color{SkyBlueCB}0.018} & 0.899\\
    & 0.05 & & 0.861 & 0.833 & 0.215 & 0.087 & 0.842 & & {\color{SkyBlueCB}\underline{0.948}} & 0.769 & {\color{SkyBlueCB}\underline{0.127}} & 0.021 & {\color{SkyBlueCB}\textbf{0.907}}\\\noalign{\smallskip}\hline\noalign{\smallskip}
  \end{tabular}
\end{table*}

\begin{table*}[t!]
\centering\footnotesize
\caption{Average fitness score of YOLO-FEDER FusionNet on datasets R1 and R2 across different bounding box loss weights. The best results are highlighted in \textbf{bold}, the second-best result is \underline{underlined}.}
  \label{tab:bboox_loss_weight_avg_fitness}
  \begin{tabular}{lcccccccccccccc}
    & \multicolumn{14}{c}{Bounding Box Regression Loss Weights}\\
    \noalign{\smallskip}\hline\noalign{\smallskip}
     & 7.50 & 6.00 & 4.50 & 3.00 & 1.50 & 1.25 & 1.00 & 0.75 & 0.50 & 0.25 & 0.20 & 0.15 & 0.10 & 0.05 \\\noalign{\smallskip}\hline\noalign{\smallskip}
    fitness~$\uparrow$ & 0.848 & 0.820 & 0.850 & 0.849 & 0.867 & 0.853 & 0.868 & 0.856 & \textbf{0.873} & 0.861 & 0.869 & 0.852 & \textbf{0.873} & \underline{0.872}
    \\\noalign{\smallskip}\hline\noalign{\smallskip}
  \end{tabular}
\end{table*}

\section{\label{supp:loss}Performance Impact of Bounding Box Loss}
In one-stage object detection frameworks such as YOLO, the loss function is typically formulated as a weighted combination of multiple objectives to enable joint optimization of object localization and classification.  Following this paradigm, the total loss $\mathcal{L}$ in YOLOv5 -- and in our baseline implementation of YOLO-FEDER FusionNet~\cite{Lenhard:2024_YOLO-FEDER} -- comprises three key components: \textit{bounding box regression loss} ($\mathcal{L}_{box}$), \textit{classification loss} ($\mathcal{L}_{cls}$), and \textit{objectness loss} ($\mathcal{L}_{obj}$). The localization term $\mathcal{L}_{box}$ is computed using the \textit{complete intersection over union (CIoU)}, while $\mathcal{L}{cls}$ and $\mathcal{L}{obj}$ are formulated using \textit{binary cross-entropy (BCE)}. Each loss component is modulated by a scalar weight to balance its contribution to the overall objective, as described below:
\begin{equation}
    \label{eq:lossv5}
    \mathcal{L} = w_1\mathcal{L}_{box} + w_2\mathcal{L}_{cls} + w_3\mathcal{L}_{obj} \ ,
\end{equation}

\noindent where $w_1=0.05$, $w_2=0.5$, and $w_3=1$. As per default configuration (cf.~\cite{Ultralytics}), the classification loss $\mathcal{L}_{cls}$ is omitted (i.e., its weight is set to zero) in single-class classification scenarios. For further implementation details, refer to~\cite{Ultralytics}. 

In anchor-free YOLO variants (v8 and beyond), the loss function was substantially revised. Most notably, the objectness loss term $\mathcal{L}_{obj}$ was replaced by the \textit{distribution focal loss} $\mathcal{L}_{dfl}$, and the relative weights of $\mathcal{L}_{box}$ and $\mathcal{L}_{cls}$ were modified to enhance detection performance on benchmark data. The resulting composite loss $\mathcal{L}$ is given by Eq.~\ref{eq:YOLOv5_loss}  (Sec.~\ref{subsec:train_details}, main paper). By default, the weighting coefficients were set to $w_1=7.5$, $w_2=0.5$, and $w_3=1.5$~\cite{Ultralytics}. As the loss function's weighting factors were empirically optimized for the COCO benchmark dataset~\cite{Lin:2014}, they may not necessarily generalize well to our specific application, network architecture, or data characteristics. To identify a more effective weighting scheme for an enhanced iteration of YOLO-FEDER FusionNet, we perform an ablation study on the loss function $\mathcal{L}$ (Eq.~\ref{eq:AFM}, main paper) as described below:\vspace{-0.1cm}

\subsection{Experimental Setup}
Loss weighting optimization is performed using an anchor-free variant of YOLO-FEDER FusionNet, which integrates a YOLOv5l backbone while maintaining the architectural structure outlined in Fig.~\ref{fig:yolofeder} (main paper). The model is trained on a hybrid dataset composed of SynDroneVision~\cite{Lenhard:2024} and DUT Anti-UAV~\cite{Zhao:2022} (cf. Sec.~\ref{subsec:data}), with training hyperparameter specified in Sec.~\ref{subsec:train_details}. To assess the impact of loss weighting on detection performance, we systematically vary the weight $w_1$ associated with the bounding box regression loss $\mathcal{L}_{\text{box}}$, while keeping the remaining weights fixed. The resulting models are evaluated on the real-world datasets R1 and R2 across multiple cropping resolutions (cf. Tab.~\ref{tab:cropping_dims}), using the evaluation metrics outlined in Sec.~\ref{subsec:eval_details}.

\subsection{Results}
Empirical analysis of the bounding box regression loss weight $w_1$ indicates that the default value proposed in~\cite{Ultralytics} is overly high, which negatively impacts detection performance (see Tab.~\ref{tab:bboox_loss_weight}). The most stable and consistently high results are observed for $w_1$ values within the range of 0.1 to 1.5, suggesting a low sensitivity to exact tuning within this interval. Based on the aggregated performance metrics and average fitness scores reported in Tab.~\ref{tab:bboox_loss_weight_avg_fitness}, a bounding box regression loss weight of $w_1 = 0.1$ appears to be a well-suited configuration for subsequent analyses.

\section{\label{supp:fitness}Sensitivity of Fitness Scores to Weight Variations}
To assess the robustness of the proposed fitness score (cf. Sec.~\ref{subsec:eval_details}, Eq.~\ref{eq:fitness}, main paper), we analyze its sensitivity to variations in performance indicator weights, exemplified by the setup of Exp.~2 (cf. Sec.~\ref{subsec:exp_details}, main paper). Tab.~\ref{tab:fitness_score} reports average fitness scores across multiple feature fusion configurations, computed as the mean over different evaluation datasets (R1 and R2) and cropping resolutions (640$\times$640 and 1080$\times$1080), under systematically varied weight settings. While the absolute magnitudes of the weights are adjusted, their relative importance is maintained across all settings.  Results show that, despite minor variations in absolute fitness scores across weighting schemes, the relative ordering of configurations -- and thus their ranking -- remains consistent (cf. Tab.~\ref{tab:fitness_score}). This indicates that the absolute scale of the weights is not critical, as long as the prioritization of the performance indicators is maintained.

\begin{table}[ht!]
\centering\footnotesize
\caption{Impact of performance indicator weight variations -- under fixed prioritization -- on average fitness scores across different feature fusion configurations (based on Exp.~2, Sec.~\ref{subsec:exp_details}, main paper).}
  \label{tab:fitness_score}
  \begin{tabular}{ccccccccccc}
  \noalign{\smallskip}\hline\noalign{\smallskip}
  \multicolumn{4}{c}{Performance Indicator Weights} & & \multicolumn{6}{c}{Average Fitness Scores by Configuration}\\
  mAP@0.25 & mAP@0.5 & FNR & FDR & & Conf.~1 & Conf.~2 & Conf.~3 & Conf.~4 & Conf.~5 & Conf.~6 \\\noalign{\smallskip}\hline\noalign{\smallskip}
    0.025 & 0.025 & 0.50 & 0.45 & & 0.880 & 0.863 & 0.866 & 0.889 & 0.868 & 0.851\\\noalign{\smallskip}\hline\noalign{\smallskip}

    0.05 & 0.05 & 0.50 & 0.40 & & 0.876 & 0.860 & 0.862 & 0.885 & 0.865 & 0.848\\\noalign{\smallskip}\hline\noalign{\smallskip}
    
    0.10 & 0.10 & 0.45 & 0.35 & & 0.873 & 0.859 & 0.859 & 0.882 & 0.861 & 0.845\\\noalign{\smallskip}\hline\noalign{\smallskip} 

    0.15 & 0.15 & 0.40 & 0.30 & & 0.870 & 0.858 & 0.857 & 0.879 & 0.857 & 0.843\\\noalign{\smallskip}\hline\noalign{\smallskip}

    0.20 & 0.20 & 0.35 & 0.25 & & 0.867 & 0.857 & 0.855 & 0.876 & 0.853 & 0.841\\\noalign{\smallskip}\hline\noalign{\smallskip}
  \end{tabular}
\end{table}

\section{\label{supp:performance_evolution}Performance Evolution of YOLO-FEDER FusionNet}
Tab.~\ref{tab:Total_Improvements} presents the progressive performance improvements of the most promising YOLO-FEDER FusionNet configuration (cf. Sec.~\ref{sec:results}, main paper) relative to its baseline~\cite{Lenhard:2024}. Specifically, it quantifies the individual contributions of (i)~optimized training data (SynDroneVision~\cite{Lenhard:2024} + DUT Anti-UAV~\cite{Zhao:2022}), (ii)~advanced feature fusion (Config. 4), and (iii) an upgraded YOLO backbone (YOLOv8l). The resulting impact on key performance metrics is reported for evaluation datasets R1 and R2 (cf. Sec.~\ref{subsec:data}, main paper) across two input resolutions (1080$\times$1080 and 640$\times$640). Each row reflects the relative change introduced by an additional modification with respect to the preceding configuration. For instance, on Dataset R2 (640$\times$640), optimizing the training data yields a relative mAP@0.5 improvement of $+$0.492 over the baseline, while incorporating DWD-based FEDER features provides an additional $+$0.142 gain (cf. Tab.~\ref{tab:Total_Improvements}, 640$\times$640, rows 1--3). Thus, these step-wise deltas highlight the cumulative contribution of each network adaption. The final row for each cropping resolution reports the aggregate improvement over the baseline.

\begin{table}[t!]
\centering\footnotesize
\caption{Progressive performance gains of YOLO-FEDER FusionNet, from the baseline~\cite{Lenhard:2024_YOLO-FEDER} to the most promising configuration (YOLOv8l backbone, DWD-based FEDER features, and optimized training data). Baseline performance is reported in absolute values across key metrics. Each subsequent row reflects the cumulative impact of successive modifications -- namely training data optimization (SynDroneVision + DUT-Anti-UAV), DWD-based feature fusion (Config. 4), and the backbone upgrade to YOLOv8l. Improvements are reported as incremental changes relative to the previous configuration. Performance gains are highlighted in green; degradations are marked in red.}
  \label{tab:Total_Improvements}
  \begin{tabular}{cccrrrrrrrrrrr}
  
 & & &\multicolumn{4}{c}{Dataset R1} & & \multicolumn{4}{c}{Dataset R2}\\
  \noalign{\smallskip}\hline\noalign{\smallskip}
  
     Img Size & Adaptions & &\multicolumn{2}{c}{ \ \ \ \ mAP~$\uparrow$} & FNR~$\downarrow$ & FDR~$\downarrow$ & & \multicolumn{2}{c}{ \ \ \ \ mAP~$\uparrow$} & FNR~$\downarrow$ & FDR~$\downarrow$\\
    & & & @0.25 & @0.5 & & & &@0.25 & @0.5\\\noalign{\smallskip}\hline\noalign{\smallskip}

    \multirow{8}*{1080$\times$1080} & (baseline) & & 0.708 & 0.636 & 0.449 & 0.066 & & 0.816 & 0.423 & 0.335 & 0.007\\\noalign{\smallskip}\cline{2-12}\noalign{\smallskip}
    & \multicolumn{1}{l}{+ Optimized Training Data} & & \color{Green}{$+$0.140} & \color{Green}{$+$0.196} & \color{Green}{$-$0.244} & \color{Red}{$+$0.100} & & \color{Green}{$+$0.130} & \color{Green}{$+$0.351} & \color{Green}{$-$0.196} & \color{Red}{$+$0.003} \\\noalign{\smallskip}\cline{2-12}\noalign{\smallskip}
    
    & \multicolumn{1}{l}{+ Feature Fusion Config. 4} & &
    \color{Green}{$+$0.005} & \color{Green}{$+$0.003} & \color{Red}{$+$0.015} & \color{Green}{$-$0.072} 
    & & \color{Green}{$+$0.016} & \color{Green}{$+$0.054}  & \color{Green}{$-$0.047}  & \color{Red}{$+$0.003} 
    \\\noalign{\smallskip}\cline{2-12}\noalign{\smallskip}
    
    & \multicolumn{1}{l}{+ YOLOv8l Backbone} & & \color{Green}{$+$0.019} & \color{Green}{$+$0.017} & \color{Green}{$-$0.028} & \color{Green}{$-$0.045}
    & & \color{Green}{$-$0.005} & \color{Red}{$-$0.018}& \color{Red}{$+$0.015} & \color{Green}{$-$0.008}

    \\\noalign{\smallskip}\cline{2-12}\noalign{\smallskip}
    & \multicolumn{1}{l}{\textbf{Total Improvement}} & & \textbf{\color{Green}{$+$0.164}} & \textbf{\color{Green}{$+$0.216}} & \textbf{\color{Green}{$-$0.257}} & \textbf{\color{Green}{$-$0.017}} & & \textbf{\color{Green}{$+$0.151}} & \textbf{\color{Green}{$+$0.387}} & \textbf{\color{Green}{$-$0.228}} & \textbf{\color{Green}{$-$0.002}}
    \\\noalign{\smallskip}\hline\noalign{\smallskip}
    
    %--------------------------------
    
    \multirow{8}*{640$\times$640} & (baseline) & & 0.729 & 0.669 &0.372 & 0.114 & & 0.685 & 0.270 & 0.473 & 0.029\\\noalign{\smallskip}\cline{2-12}\noalign{\smallskip}
    
    & \multicolumn{1}{l}{+ Optimized Training Data} & &  \color{Green}{$+$0.150} & \color{Green}{$+$0.183} & \color{Green}{$-$0.175} & \color{Green}{$-$0.069} &  & \color{Green}{$+$0.261} & \color{Green}{$+$0.492} & \color{Green}{$-$0.327} & \color{Green}{$-$0.011}
    \\\noalign{\smallskip}\cline{2-12}\noalign{\smallskip}
    
    & \multicolumn{1}{l}{+ Feature Fusion Config. 4} & & \color{Red}{$-$0.026} & 
    \color{Red}{$-$0.022} & \color{Red}{$+$0.032} & \color{Red}{$+$0.010} 
    & & \color{Red}{$-$0.002} & \color{Green}{$+$0.142} & \color{Green}{$-$0.025} & \color{Red}{$+$0.008}
    \\\noalign{\smallskip}\cline{2-12}\noalign{\smallskip}
    
    & \multicolumn{1}{l}{+ YOLOv8l Backbone} & &
    
    \color{Green}{$+$0.023}& \color{Green}{$+$0.018} & \color{Green}{$-$0.041} & \color{Red}{$+$0.061}
    & & \color{Green}{$+$0.023} & \color{Red}{$-$0.006} & \color{Green}{$-$0.039} & \color{Green}{$-$0.007}
    \\\noalign{\smallskip}\cline{2-12}\noalign{\smallskip}
    
    & \multicolumn{1}{l}{\textbf{Total Improvement}} & & \textbf{\color{Green}{$+$0.147}} & \textbf{\color{Green}{$+$0.179}} & \textbf{\color{Green}{$-$0.184}} & \textbf{\color{Red}{$+$0.002}} & & \textbf{\color{Green}{$+$0.282}} & \textbf{\color{Green}{$+$0.628}} & \textbf{\color{Green}{$-$0.391}} & \textbf{\color{Green}{$-$0.010}}
    \\\noalign{\smallskip}\hline\noalign{\smallskip}
    
  \end{tabular}
\end{table}

\vfill
\end{document}